\documentclass[conference,compsoc]{IEEEtran}
\IEEEoverridecommandlockouts
\def\BibTeX{{\rm B\kern-.05em{\sc i\kern-.025em b}\kern-.08emT\kern-.1667em\lower.7ex\hbox{E}\kern-.125emX}}

\usepackage[
  style=ieee,
  citestyle=numeric-comp,
  maxbibnames=6,
  minbibnames=1,
  doi=false,
  isbn=false,
  url=false,
  eprint=false,
]{biblatex}

\AtEveryBibitem{%
  \clearfield{month}%
  \clearfield{day}%
  \clearfield{location}%
  \clearfield{address}%
  \clearfield{publisher}%
  \clearfield{note}%
  \clearlist{language}%
  \clearfield{volume}%
  \clearfield{number}%
  \clearfield{pages}%
  \clearfield{organization}%
}

\AtBeginBibliography{%
  \renewbibmacro*{doi+eprint+url}{}
}

\usepackage{nicefrac}
\usepackage{siunitx}
\usepackage{array,framed}
\usepackage{booktabs}
\usepackage{
  color,
  float,
  epsfig,
  wrapfig,
  graphics,
  graphicx,
  subcaption
}
\usepackage{textcomp}
\usepackage{setspace}
\usepackage{amsfonts}
\usepackage{latexsym,fancyhdr,url}
\usepackage{enumerate}
\usepackage{algorithm}
\usepackage{algpseudocode}
\usepackage{graphics}
\usepackage{xparse}
\usepackage{xspace}
\usepackage{multirow}
\usepackage{csvsimple}
\usepackage{balance}
\usepackage{setspace}
\usepackage{threeparttable}
\usepackage{pifont}

\def\ie{{i.e.},~}
\def\eg{{e.g.},~}
\usepackage{microtype}

\usepackage{
  tikz,
  pgfplots,
  pgfplotstable
}
\usepackage[hidelinks]{hyperref} 
\usetikzlibrary{
  shapes.geometric,
  arrows,
  external,
  pgfplots.groupplots,
  matrix
}
\pgfplotsset{compat=1.9}

\usepackage{mathtools,}

\DeclareMathAlphabet{\mathcal}{OMS}{cmsy}{m}{n}
\DeclareGraphicsExtensions{%
    .png,.PNG,%
    .pdf,.PDF,%
    .jpg,.mps,.jpeg,.jbig2,.jb2,.JPG,.JPEG,.JBIG2,.JB2}

\usepackage{xparse}
\newcommand{\bnm}{\begin{newmath}}
\newcommand{\enm}{\end{newmath}}

\newcommand{\bea}{\begin{eqnarray*}}%
\newcommand{\eea}{\end{eqnarray*}}%

\newcommand{\bne}{\begin{newequation}}
\newcommand{\ene}{\end{newequation}}

\newcommand{\bal}{\begin{newalign}}
\newcommand{\eal}{\end{newalign}}

\newenvironment{newalign}{\begin{align}%
\setlength{\abovedisplayskip}{4pt}%
\setlength{\belowdisplayskip}{4pt}%
\setlength{\abovedisplayshortskip}{6pt}%
\setlength{\belowdisplayshortskip}{6pt} }{\end{align}}

\newenvironment{newmath}{\begin{displaymath}%
\setlength{\abovedisplayskip}{4pt}%
\setlength{\belowdisplayskip}{4pt}%
\setlength{\abovedisplayshortskip}{6pt}%
\setlength{\belowdisplayshortskip}{6pt} }{\end{displaymath}}

\newenvironment{newequation}{\begin{equation}%
\setlength{\abovedisplayskip}{4pt}%
\setlength{\belowdisplayskip}{4pt}%
\setlength{\abovedisplayshortskip}{6pt}%
\setlength{\belowdisplayshortskip}{6pt} }{\end{equation}}

\newcounter{ctr}

\newcounter{mytable}
\def\mytable{\begin{centering}\refstepcounter{mytable}}
\def\endmytable{\end{centering}}

\newcounter{myfig}
\def\myfig{\begin{centering}\refstepcounter{myfig}}
\def\endmyfig{\end{centering}}

\newlength{\saveparindent}
\newlength{\saveparskip}
\newcommand{\E}{{\rm I\kern-.3em E}}

\renewcommand{\eqref}[1]{\mbox{Equation~(\ref{#1})}}

\def \part {part}

\renewcommand{\paragraph}[1]{\vspace*{6pt}\noindent\textbf{#1}\;}

\def \blackslug{\hbox{\hskip 1pt \vrule width 4pt height 8pt
    depth 1.5pt \hskip 1pt}}
\def \qed{\quad\blackslug\lower 8.5pt\null\par}

\newcounter{mynote}[section]

\newcommand\ignore[1]{}

\newcounter{rcnote}[section]

\newcounter{mrnote}[section]

\newcounter{fknote}[section]

\newcounter{anote}[section]

\DeclareMathSymbol{\mlq}{\mathord}{operators}{``}
\DeclareMathSymbol{\mrq}{\mathord}{operators}{`'}

\newcommand{\rhf}[2]{R_{f, \gamma}}

\DeclareDocumentCommand{\edist}{o o}{
  \ensuremath{
    \IfNoValueTF{#1}{{d}}{{\sf d}(#1,#2)}
  }
}

\newcommand{\olrk}[1]{\ifx\nursymbol#1\else\!\!\mskip4.5mu plus 0.5mu\left(\mskip0.5mu plus0.5mu #1\mskip1.5mu plus0.5mu \right)\fi}

\NewDocumentCommand{\indseq}{ O{1} O{r} }{{#1}\ldots {#2}}

\DeclareRobustCommand*\circled[1]{\tikz[baseline=(char.base)]{\node[shape=circle,draw,color=white,fill=black,inner sep=0.5pt] (char){#1};}}

\pgfplotsset{compat=1.11,
    /pgfplots/ybar legend/.style={
    /pgfplots/legend image code/.code={%
       \draw[##1,/tikz/.cd,yshift=-0.25em]
        (0cm,0cm) rectangle (3pt,0.8em);},
   },
}

\usepackage{comment}

\algrenewcommand\algorithmicrequire{\textbf{Input:}}
\algrenewcommand\algorithmicensure{\textbf{Output:}}
\usepackage{xspace}
\newcommand{\AdvTraj}{{\textsc{\small{AdvTraj}}}\xspace}
\usepackage[scaled=.75]{beramono}

\newcommand{\IDa}{\mathtt{ID_a}}
\newcommand{\IDb}{\mathtt{ID_b}}

\makeatletter
\newcommand{\linebreakand}{%
  \end{@IEEEauthorhalign}
  \hfill\mbox{}\par
  \mbox{}\hfill\begin{@IEEEauthorhalign}
}
\makeatother

\begin{document}

\title{Physical ID-Transfer Attacks against Multi-Object Tracking via \\ Adversarial Trajectory\\
}

\author{\IEEEauthorblockN{Chenyi Wang}
\IEEEauthorblockA{\textit{University of Arizona} \\
chenyiw@arizona.edu}
\and
\IEEEauthorblockN{Yanmao Man}
\IEEEauthorblockA{\textit{HERE Technologies} \\
yman@arizona.edu}
\and
\IEEEauthorblockN{Raymond Muller}
\IEEEauthorblockA{\textit{Purdue University} \\
mullerr@purdue.edu}
\and
\IEEEauthorblockN{Ming Li}
\IEEEauthorblockA{\textit{University of Arizona} \\
lim@arizona.edu}
\linebreakand
\IEEEauthorblockN{Z. Berkay Celik}
\IEEEauthorblockA{\textit{Purdue University} \\
zcelik@purdue.edu}
\and
\IEEEauthorblockN{Ryan Gerdes}
\IEEEauthorblockA{\textit{Virginia Tech} \\
rgerdes@vt.edu}
\and
\IEEEauthorblockN{Jonathan Petit}
\IEEEauthorblockA{\textit{Qualcomm} \\
petit@qti.qualcomm.com}
}

\date{}

\maketitle

\begin{abstract}
Multi-Object Tracking (MOT) is a critical task in computer vision, with applications ranging from surveillance systems to autonomous driving. However, threats to MOT algorithms have yet been widely studied. In particular, incorrect association between the tracked objects and their assigned IDs can lead to severe consequences, such as wrong trajectory predictions. Previous attacks against MOT either focused on hijacking the trackers of individual objects, or manipulating the tracker IDs in MOT by attacking the integrated object detection (OD) module in the digital domain, which are model-specific, non-robust, and only able to affect specific samples in offline datasets. In this paper, we present \AdvTraj, the first online and physical ID-manipulation attack against tracking-by-detection MOT, in which an attacker uses adversarial trajectories to transfer its ID to a targeted object to confuse the tracking system, without attacking OD. Our simulation results in CARLA show that \AdvTraj can fool ID assignments with 100\% success rate in various scenarios for white-box attacks against SORT, which also have high attack transferability (up to 93\% attack success rate) against state-of-the-art (SOTA) MOT algorithms due to their common design principles. We characterize the patterns of trajectories generated by \AdvTraj and propose two universal adversarial maneuvers that can be performed by a human walker/driver in daily scenarios. Our work reveals under-explored weaknesses in the object association phase of SOTA MOT systems, and provides insights into enhancing the robustness of such systems.\looseness=-1

\end{abstract}

\pagenumbering{arabic}

\begin{figure}[t]
    \centering
    \includegraphics[width=1.0\linewidth]{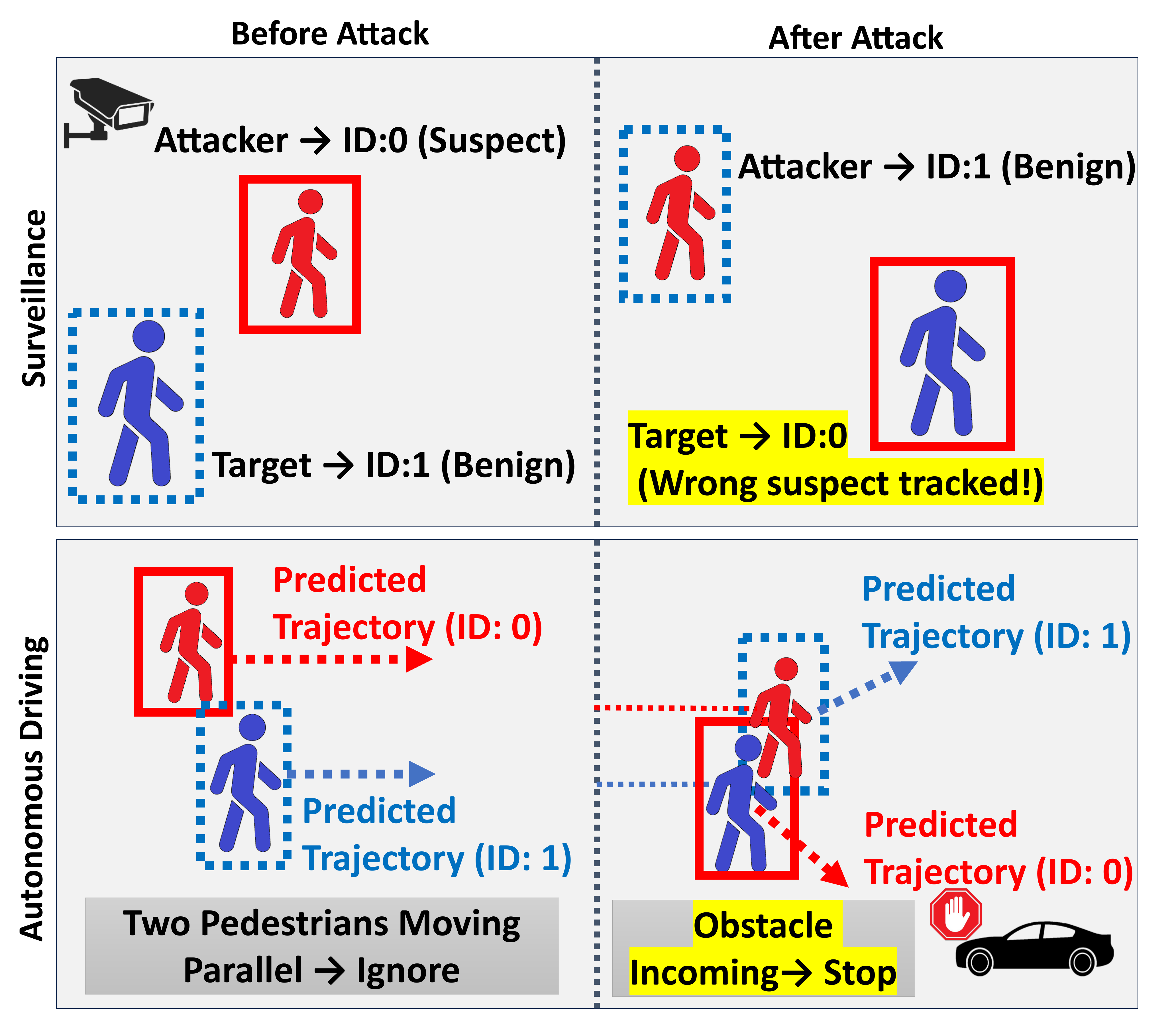}
    \caption{Illustration of potential consequences of ID-Transfer in surveillance and autonomous driving (AD) applications. In the surveillance scenario (above), ID-Transfer leads to wrong target of interest being tracked. In the AD scenario (below), ID-Transfer results in inaccurate trajectory prediction due to history trajectories that are inconsistent with ground truths.\looseness=-1
   }
    \label{fig:motivation}
\end{figure}

\section{Introduction}

In computer vision, Multi-Object Tracking (MOT) algorithms play a pivotal role in understanding and interpreting dynamic scenes. These algorithms are designed to track multiple objects simultaneously by assigning unique IDs as they move across video frames. With applications ranging from autonomous driving (AD)~\cite{BaiduApollo, Waymo, OpenPilot, Autoware} and pedestrian/vehicle surveillance systems~\cite{elhoseny2020surveillance} to military unmanned aerial vehicles~\cite{militaryUAV1, militaryUAV2}, the assigned IDs of the MOT system are used to uniquely identify objects of interest for trajectory predictions in AD and suspect tracking in surveillance systems. Due to the safety-critical nature of these applications, the correct and consistent association between assigned IDs and tracked objects is of crucial importance. For example, in surveillance systems, as shown in
Figure~\ref{fig:motivation}, accurate and consistent identification allows
effective monitoring and timely response to incidents.
An ID mismatch
can lead to losing track of the object of interest, resulting in wrongful
accusations or the escape of tracking by a criminal. 
Furthermore, AD systems (\eg Baidu Apollo~\cite{BaiduApollo}) typically operate through a pipeline that includes perception, object tracking, trajectory prediction, planning, and control. If tracked objects' IDs are mismatched, the prediction module will make wrong trajectory predictions based on incorrect trajectory histories.

\begin{figure*}[t]
\centering
\includegraphics[width=0.9\textwidth]{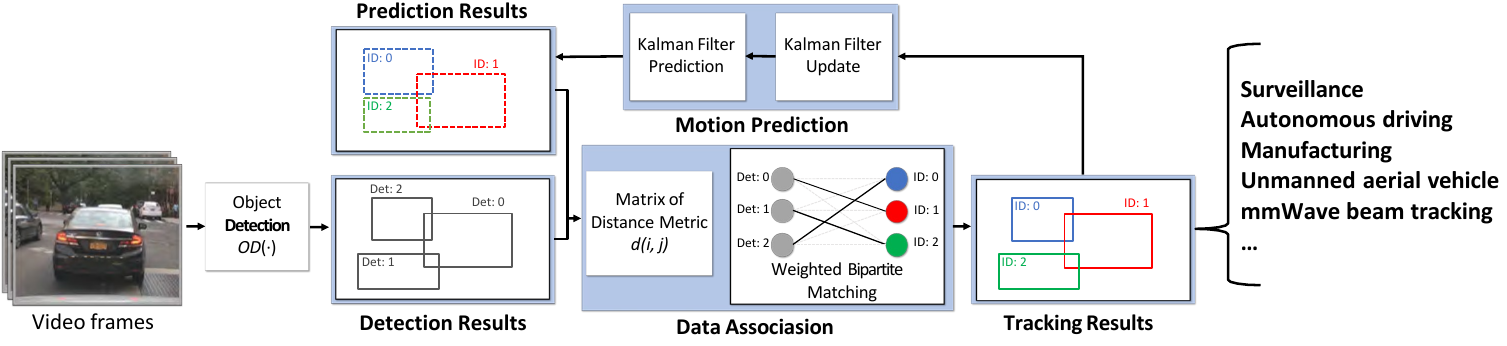}
\caption{Illustration of tracking-by-detection pipeline in Multi-Object Tracking (MOT).}
\label{fig:tracking by detection}
\end{figure*}

Most state-of-the-art (SOTA) MOT algorithms (\eg ByteTrack~\cite{ByteTrack}, OC-SORT~\cite{OCSORT}, etc.) generally follow the tracking-by-detection paradigm. This consists of two explicit stages: ($1$) an object detection (OD) phase that produces bounding box detections, and ($2$) an association phase that performs weighted bipartite matching between detections and bounding boxes produced by motion prediction models. The motion prediction model, such as Kalman Filter~\cite{kalman1960new}, maintained by individually indexed trackers, is an integrated and necessary component of MOT. This is because other information that can potentially be used for object identification (such as facial recognition or other biometric features) is not always available due to privacy concerns, different camera angles, low resolution, or different application scenarios (\eg vehicle tracking), etc.\looseness=-1

Although previous works have shown that MOT is vulnerable to several types of attacks,  they have mainly targeted the OD component, and primarily aimed at inducing errors in an individual object's detected trajectory. For example, recent tracker hijacking attacks~\cite{Jia2020TrackerHijack, Ma2023Attack} cause the detected trajectory to deviate from the true track of an object. Other works investigated identity-switching attacks that cause a target object's ID to be switched to a different one~\cite{Lin2021TraSw, FFAttack}, causing a loss of tracking on the target of interest. 
However, these works are highly dependent on traditional perturbation-based attacks against OD
that require strong attacker capability, such as full read and write access to
the video feed in an offline dataset and white-box knowledge of the OD model
structure and parameters,
which limits their practical impact.\looseness=-1

In this paper, we introduce \AdvTraj, a novel physical and online ID-Transfer attack that confuses the tracking of \emph{two} objects, rather than introducing a tracking error on an individual object. Instead of attacking the OD models of MOT, our attack exploits the vulnerability of the association phase (especially motion prediction), by using physically realizable yet adversarial trajectories to confuse the ID assignment and matching algorithm. 
Unlike previous works, we consider a stealthier ID-manipulation attack in which the attacker
aims to \emph{transfer} its MOT-assigned ID to another targeted and
non-cooperative object without losing the attacker's original track ID,
because otherwise it could raise suspicion. In addition, we consider a more
realistic threat model where the attack is online, physically realizable, and
does not require digital modification of the video input.\looseness=-1

We start by assuming an adversary with white-box knowledge of the MOT algorithm. 
By deriving conditions on desired ID assignments and optimizing for physical trajectories, \AdvTraj addresses several technical challenges: ~(a) non-differentiability in the bipartite matching algorithm, (b) manipulating detected bounding boxes, and (c) adherence to physical constraints. 
Hence, the attack can be conducted in real-time where the perceived image sequence of the MOT system is genuine, but contains the adversarial trajectory. 
We implement our white-box attack against the SORT algorithm~\cite{SORT}, and then show it can be readily transferable to other SOTA MOT algorithms under the black-box setting, due to their common design principles.

By investigating the patterns of adversarial
trajectories generated by \AdvTraj, we identify several underlying
characteristics of these trajectories. Based on this, we further develop universal adversarial maneuvers (UAMs) that can be easily realized by a human walker/driver, since executing optimized adversarial trajectory in the real world would require precise motor control. 
These UAMs effectively introduce
discrepancies between the attacker's actual and predicted locations when being
close to the target, by performing a nonlinear movement (\eg acceleration and
deceleration), which induces ID-Transfer. Our contributions are summarized as follows:
\looseness=-1


\begin{itemize}

\item We introduce \AdvTraj, the \emph{first} online and physical ID-Transfer attack against MOT algorithms using an adversarial trajectory. \AdvTraj enables an attacker to disrupt the MOT's ID assignment by transferring its own ID to another target, manipulating the system's ability to correctly track objects.

\item We show that the white-box attack against SORT is also transferable to other SOTA MOT algorithms, due to their common design principles, which eliminates the requirement of adversary's knowledge of the MOT algorithm.\looseness=-1

\item We evaluate \AdvTraj in CARLA for surveillance and autonomous driving applications. Our simulation evaluation demonstrates an attack success rate of 100\% for white-box attacks against SORT, and up to 93\% attack success rate for transferred black-box attacks against 5 other SOTA MOT algorithms.

\item We characterize the patterns of optimized adversarial trajectories discovered using \AdvTraj and propose two universal (heuristic) adversarial maneuvers that are easily realizable, which achieve up to 45\% attack success rates in our real-world experiments.

\item We discuss a set of potential countermeasures to mitigate the vulnerabilities in the association stage of MOT algorithms
against ID-Transfer attacks.

\end{itemize}
  
\section{Background and Related Work}\label{sec:background}

\subsection{Multi-Object Tracking}

Object tracking (OT) is a fundamental task for analyzing image sequences. Compared to object detection (OD) algorithms, which perform object classification and localization on a static image, OT extends the task to consistently identify detected objects across frames by assigning unique IDs to distinct object instances.
There are two categories for OT algorithms based on their goals.
Single-object tracking (SOT)~\cite{DBLP:conf/cvpr/FanLYCDYBXLL19,
DBLP:conf/eccv/MullerBGAG18} tracks a single object specified in the
reference frame and provides localization results in subsequent
frames. In contrast, multi-object tracking (MOT)~\cite{DeepSORT, SORT}
simultaneously matches multiple detected objects with
previous trajectories, offering extensive utilities with applications ranging
from surveillance~\cite{elhoseny2020surveillance} and autonomous driving systems~\cite{BaiduApollo, Waymo}, to unmanned aerial vehicles~\cite{militaryUAV1}.\looseness=-1

MOT algorithms can themselves be categorized into two paradigms:
joint-detection-and-tracking and tracking-by-detection. The
former is an end-to-end approach that aims to unify the
detection and tracking processes into a single cohesive model~\cite{centerTrack, retinaTrack, meinhardt2021trackformer}. On the other hand,
the tracking-by-detection framework
consists of two stages where the system identifies objects in each frame
explicitly using OD models, and then makes associations across frames to form
trajectories~\cite{DeepSORT, OCSORT, BoTSORT}, which are more commonly used
in autonomous systems~\cite{BaiduApollo,Autoware,Ma2023Attack}.

Figure~\ref{fig:tracking by detection} shows the general pipeline of tracking-by-detection. Upon arrival of each frame, the system calls the OD model to locate detected objects and passes the bounding boxes to the tracking (\ie association) module. 
For each tracked object, the tracker maintains a motion prediction model and optionally the Re-Identification (ReID) features, which use deep learning models to represent the object's appearance in past frames~\cite{CVPR16ResNet, MobileNet, ICCV19OSNet}. The motion prediction model in each tracker, usually implemented using Kalman Filter~\cite{kalman1960new}, is used to produce a predicted bounding box position based on the object's past state estimates.  Finally, the tracking module performs weighted bipartite graph matching based on common distance metrics: ($1$) Intersection-over-Union (IoU) between detected and tracker-predicted bounding boxes, and/or ($2$) distance in ReID feature space.\looseness=-1

\subsection{Existing Attacks}

Since the introduction of adversarial examples against image classification~\cite{SzegedyZSBEGF13}, various attacks have been proposed against object detection (OD)~\cite{DAGAttack, ShapeShifter, DPatch, Eykholt18, TOGAttack} and single-object tracking (SOT) ~\cite{CoolingShrinking, SparkAttack, HijackingTracker, UnivAttackSOT, muller22TrackerHijacking}. 
Although OD is an integrated and vital stage of tracking-by-detection based MOT algorithms, it has been shown that an attack targeting OD (\eg vanishing attack) needs to succeed at least 98\% of the time over 60 consecutive frames to influence the MOT algorithms~\cite{Jia2020TrackerHijack}. 
To date, no OD attack has been able to achieve such a high success rate. In addition, the real-world impact of attacks on SOT remains unclear, since most surveillance and AD systems adopt MOT. 
Although MOTs are used in safety-critical applications, there are only a limited number of MOT attacks due to the difficulty of directly applying OD attacks in the MOT pipeline. 

Depending on the primary attack objectives, current MOT attacks can be broadly divided into two categories: (a) tracker hijacking and (b) identity switching. 
Tracker hijacking attacks~\cite{Jia2020TrackerHijack, Ma2023Attack, muller22TrackerHijacking} aim to mislead the tracking of a target object to an incorrect trajectory by suppressing the true detected bounding box while fabricating fake ones at adjacent but wrong locations. 
The attack can be applied in both SOT and MOT on a single target object. On the other hand, identity switching attacks~\cite{Lin2021TraSw, FFAttack} are applied to MOT where the attacker manipulates the detected bounding boxes to switch the identity of a target object assigned by the MOT system to a different one. Although these two types of attacks differ in attack goals, they are evaluated using the same metric of ID-Switch, where an attack is considered successful if the target object switches to a new ID after the attack (while its old one is not necessarily preserved).

Jia et al.~\cite{Jia2020TrackerHijack} proposed an offline tracker hijacking attack in which an attacker places an adversarial patch to make the detected object bounding box shift toward the direction opposite to its true movement, resulting in the original tracker being misled and eventually lost while the tracked object switches to a new identity after a certain number of frames. However, the attack was only demonstrated in the digital domain where only a single tracked object is present. The attack's practicality and robustness were questioned in their subsequent work in progress~\cite{Ma2023Attack} where its adversarial patch must simultaneously achieve both removal and fabrication of the bounding box. Also, they did not evaluate the attack's transferability to more advanced MOT algorithms and the effect of digital perturbation on the ReID feature. Therefore, the effectiveness of this attack in more practical settings remains unclear.\looseness=-1

Muller et al.~\cite{muller22TrackerHijacking} introduced AttrackZone, an online and physical tracker hijacking attack against Siamese trackers~\cite{SiameseTracker} that exploits the heatmap generation process of Siamese Region Proposal Networks to take control of an object's bounding box. The attack utilizes projectors to present adversarial noise in physically dark environments, which limits its effectiveness in more general scenarios. Although being an online and robust attack against OT, AttrackZone only applies to specific SOT algorithms capable of tracking one target object, and its real-world impact on surveillance systems and autonomous vehicles that rely on MOT algorithms remains open.\looseness=-1

\begin{figure}[t]
\centering
\includegraphics[width=0.9\columnwidth]{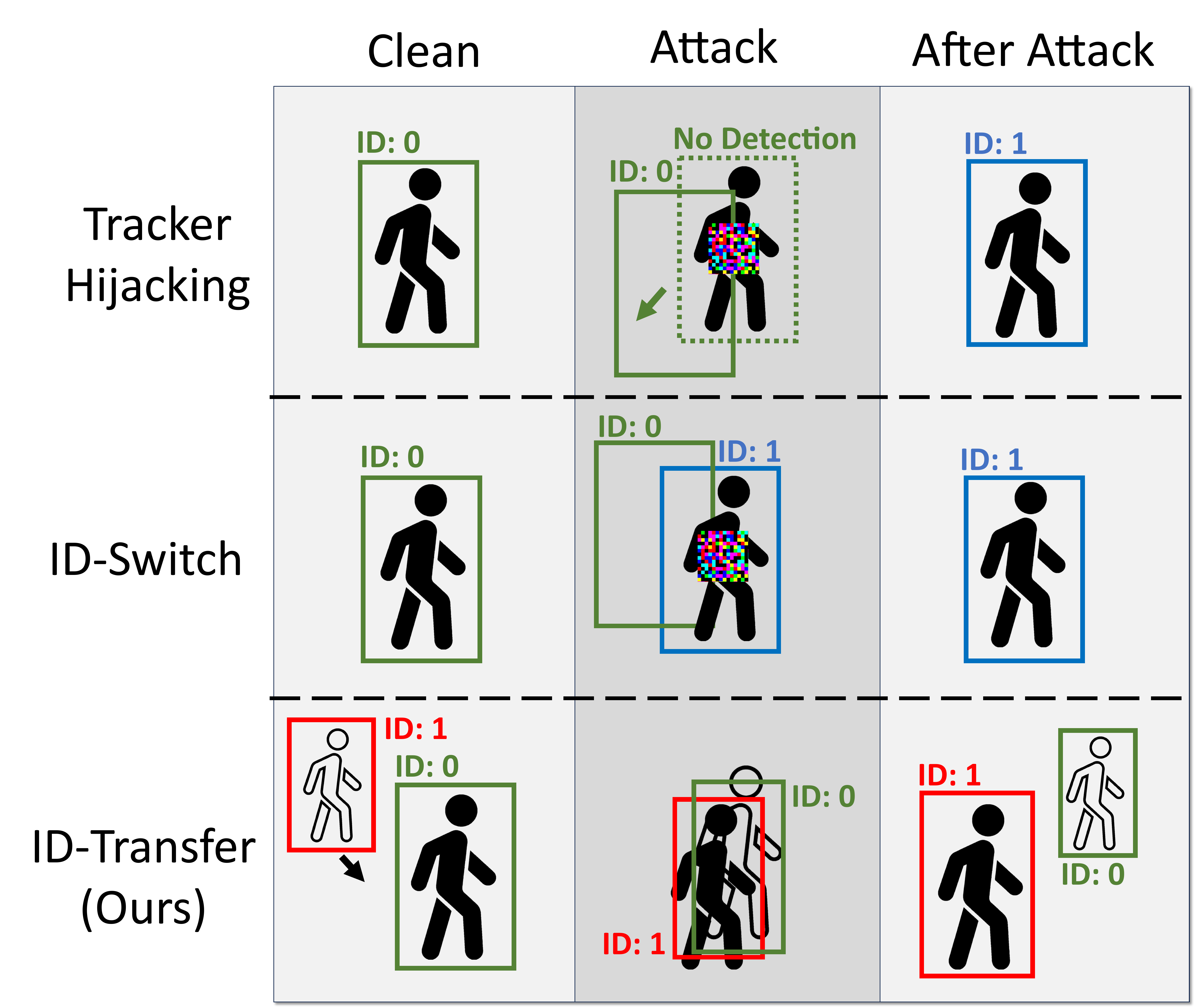}
\caption{Comparison of existing MOT attacks (by attacking OD modules in digital space) and our ID-Transfer attack (by adversarial trajectories in physical space).}
\label{fig:attack_comparison}
\end{figure}

There are two notable identity switching attacks in MOT.
Lin et al.~\cite{Lin2021TraSw} introduced the tracklet-switch attack against two specific trackers: FairMOT~\cite{FairMOT} and ByteTrack~\cite{ByteTrack}. 
They proposed a method to generate digital perturbations to make MOT algorithms confuse intersecting pedestrian trajectories in an offline setting. Although this attack also targets pedestrian tracking scenarios, it is similar to the tracker hijacking attack~\cite{Jia2020TrackerHijack} since it aims to switch the ID assignment in one of the two pedestrians to a different value (not necessarily preserving any original IDs). 
In addition, they assumed that the attacker can arbitrarily manipulate the captured video frame to add perturbations that affect the bounding box positions and ReID features of multiple objects at the same time. Such a strong threat model limits the practicality of this attack.\looseness=-1

Another work introduced the False-Positive-and-False-Negative attack~\cite{FFAttack} that creates a new identity for a targeted object by erasing the true bounding box while fabricating multiple similar-sized bounding boxes around it. 
The fake detections, assigned with identities different from the original, will form random trajectories around the true object. 
After the attack, when the true bounding box detection is restored, it is assigned to one of the new trackers initiated for the fake bounding boxes, resulting in a switch in identity. However, randomly fabricated bounding boxes can cause spatio-temporal inconsistencies that may be detected by existing consistency-based defense methods (\eg\cite{PercepGuard, Muller_VOGUES}). Furthermore, the attack only works in the digital domain and is shown to be ineffective against advanced MOT algorithms with ReID models enabled~\cite{FFAttack}.\looseness=-1

All of these attacks rely on the successful execution of digital perturbations against specific OD models to manipulate the detected bounding boxes (\eg suppression and fabrication). Therefore, established defenses against OD attacks~\cite{AdvTrainingOD, PixelDetection, JPEGCompression, GaborConv} can also mitigate the existing MOT attacks. 
Meanwhile, the offline setting for these attacks and the dependence on suitable samples also raise questions about their utility in the physical domain. 
Inspired by adversarial trajectory/maneuver attacks against AD systems~\cite{Qi_adversarial_traj, acero}, which explore subtle manipulation of a vehicle's movement pattern to compromise an AD's trajectory prediction or decision-making process, our ID-Transfer attack against MOT systems
leverages physically realizable trajectory generated in an online manner without fooling the perception module. 
The input to the perception module is unaltered, eliminating the need to attack the OD model or add adversarial perturbations to the input image stream. Thus, our attack
fundamentally evades existing OD defenses against perturbation-based attacks. Figure~\ref{fig:attack_comparison} visually compares the different attack goals and methods between existing MOT attacks and our ID-Transfer attack.\looseness=-1

\section{Problem Statement}\label{sec:system}

\subsection{System Model}

We consider a real-time tracking-by-detection MOT system that is deployed in pedestrian/vehicle surveillance~\cite{BaiduApollo, elhoseny2020surveillance} or autonomous driving (AD)~\cite{BaiduApollo} systems for perception. At each time step, the MOT algorithm takes as input a  detection state vector $\mathbf{d}_i=(x_1,y_1,x_2,y_2)_i$ representing the upper-left and lower-right corner of a bounding box, for each object $O_i\in \{O_1,...,O_n\}$ from an object detection (OD) module. The OD module takes input from an RGB camera
and is assumed to be capable of providing accurate bounding box prediction for each object of interest in the scene. This assumption isolates the MOT system's ID assignment results to be solely dependent on the object trajectories (which we study in this work) where any potential changes in ID assignment results shall not be attributed to detection errors (\eg missing detection due to occlusions).\looseness=-1

The MOT system maintains a pool of trackers $\mathcal{T}=\{T_1,...,T_n\}$ for tracked objects. Each tracker keeps track of the associated object states $\mathbf{x}=(u,v,s,r,\Vec{u},\Vec{v},\Vec{s})^T\in\mathcal{X}$, where $\Vec{\cdot}$ denotes the first-order derivative, and $(u,v),s,r$ are the bounding box center position, scale and aspect ratio, respectively. The tracker is capable of predicting the states $\hat{\mathbf{x}}_\text{prior}^t$ of the associated object for the current time step $t$, which is commonly achieved using the Kalman Filter (KF)~\cite{kalman1960new}. Each tracker's KF makes such predictions by applying a state transition matrix $\mathbf{F}$ to the previous state estimates assuming linear movement between each time step~\cite{SORT}:\looseness=-1
{
\begin{equation}\label{eq:kf_pred}
\hat{\mathbf{x}}_\text{prior}^t=\mathbf{F}\hat{\mathbf{x}}_\text{posterior}^{t-1}
\end{equation}
}

The predicted states 
of the trackers are used to compare with the OD results for the association. This is achieved by calculating distance metrics based on Intersection-over-Union (IoU) or its variants $d:\mathbb{R}^4\times \mathbb{R}^4\to [0,1]$ between the detected and predicted bounding boxes for each detection-tracker pair $(i,j)$ to form the distance matrix~\cite{zheng2020diou}.
The MOT system then solves the weighted bipartite matching problem between existing trackers $T_j\in \{T_1,...,T_m\}$ and object detections $\mathbf{d}_i\in \{\mathbf{d}_1,...,\mathbf{d}_n\}$ that minimizes the metric sum using the Hungarian Algorithm~\cite{hungarian}. 
We say that the MOT system assigns ID $j$ (tracker $T_j$) to an object $O_i$ if and only if there is an edge between $(\mathbf{d}_i, T_j)$ as indicated in the assignment matrix. For brevity, we denote $f(O^t_i)=j$ if the detection of object $O_i$ is assigned to the tracker $T_j$ at time $t$. 
After the association phase, each object's detection result is used to perform the KF state update step on the tracker to which it is assigned and obtain the posterior state estimates $\hat{\mathbf{x}}_\text{posterior}^{t}$. The posterior estimates are then used to calculate the prior state estimates for the next time step $t+1$. More details about the matching process can be found in Appendix~\ref{app:bipartite_matching}. \looseness=-1

\subsection{Problem Formulation}

Initially, at $t=0$, the attacker $O_a$ is assigned to a tracker $T_{\IDa}$ while a targeted object $O_b$ is assigned to a tracker $T_{\IDb}$. The ID-Transfer attack is defined as $f(O_a^t)=\IDa,f(O_b^t)=\IDb,\forall t<\tau$ while $f(O_a^t)\neq \IDa,f(O_b^t)=\IDa, \forall t\geq \tau$. In other words, starting at time $t=\tau$, the attacker switches to a different ID, while its original ID is transferred to the target, as assigned by the MOT system. The attacker aims to craft a series of inputs into the MOT system to achieve this objective.\looseness=-1

\subsection{Threat Model}

We consider an attacker acting as a physical entity that is visible to and tracked by the MOT system. We assume that the attacker and target belong to the same class (\eg both are pedestrians or vehicles) so that they have comparable sizes. We assume the attacker to be aware of the victim MOT system's camera location and angle, and can observe the trajectory of the target object. The attacker aims to change its assigned MOT ID to a new one while transferring its original ID to the target object tracked by the system. 
The adversary approaches the above attack goal in the physical world by maneuvering itself along a physically realizable \emph{adversarial trajectory} represented by a series of waypoints (geographical locations in the real world) that will lead to the desired consequences.

The adversary's capability generally fall into two categories: $(\mathtt{T1})$ Optimized adversary (\eg autonomous vehicles or robots), who  possess the capability for real-time calculation and precise motor control, and $(\mathtt{T2})$ Heuristic adversary (\eg pedestrians or human drivers), who can only perform inexact maneuvers that rely on instinct and heuristics.  In addition, we consider  adversaries with different knowledge levels of the MOT system:

\vspace{2pt}\noindent\textbf{White-box Attacker.} We consider a white-box attacker who has full knowledge of the MOT algorithm. 
White-box attacks aim to generate optimized trajectories online as the target moves, which requires $(\mathtt{T1})$ adversary.

\vspace{2pt}\noindent\textbf{Black-box Attacker.} We consider a black-box attacker with no knowledge of the MOT algorithm being deployed in the system. The attack can be performed by either $(\mathtt{T1})$ or $(\mathtt{T2})$ adversary. The $(\mathtt{T1})$ adversary performs transfer attacks against the system by executing \AdvTraj on a surrogate model such as SORT~\cite{SORT}, while the $(\mathtt{T2})$ adversary executes adversarial maneuvers based on heuristics.

Both types of attackers are assumed to have realistically limited physical maneuverability. For example, an adversarial walker can only move at reasonable speeds ($\eg$ 0-3 m/s).
\section{\AdvTraj: Online ID-Transfer Attack}\label{sec:attack}

In this section, we outline the unique challenges of achieving the ID-Transfer attack and our solutions. We further illustrate the conditions necessary to achieve ID-Transfer in the association module of MOT algorithms, which motivates our design of \AdvTraj. 
Our attack pipeline is plug-and-play for tracking-by-detection MOT algorithms that take into account motion information, by employing the corresponding distance metrics into the loss functions.\looseness=-1 

Furthermore, we summarize the patterns of generated adversarial trajectories under random initial conditions (relative starting positions of the attacker and target) and categorize them into two base cases which other situations can reduce to. Based on this, we develop two highly executable universal adversarial maneuvers that exploit the fundamental vulnerability of the MOT algorithms.\looseness=-1

\subsection{Challenges}

\vspace{2pt}\noindent\textbf{Loss Function Design.} The non-differentiability in the matching phase of MOT and limited attacker capabilities raise two design challenges for the loss function. First, the weighted bipartite matching algorithm in the association module of MOT is discrete and hence non-differentiable. 
Thus, we derive sufficient conditions on the bounding box input to the association module that leads to the desired ID assignments, and perform optimizations with respect to 
the intermediate results of MOT consisting of the OD detections and KF predictions.
Second, our threat model assumes the attacker can only control its own physical movement without the capability to tamper with other objects' detection or motion prediction. It is challenging for the attacker to ``prescribe'' its own ID to the target under this limited capability assumption, since bipartite matching optimizes for the \emph{sum of distance metrics}. Specifically, only guiding the attacker tracker's KF-predicted states to match the target's detection does not warrant ID-Transfer since the sum of distance metrics for correct ID assignment can still dominate. Thus, we design the adversarial loss function to achieve two objectives simultaneously: match the attacker's OD detection to the target's tracker prediction and match the attacker's tracker prediction to the target's OD detection, while the attacker only controls its own movement.\looseness=-1

\vspace{2pt}\noindent\textbf{Manipulating Detected Bounding Boxes.} The attacker needs to effectively manipulate the detected bounding box sequence to affect the matching algorithm output in a controlled manner. Contrary to existing MOT attacks that focused on attacking the OD module, which have been shown to be non-robust and model-specific~\cite{Ma2023Attack}, our approach uses physical movement to create an \emph{genuine but adversarial} bounding box sequence. This strategy ensures robustness and independence from specific OD models and also evades OD defenses. 
Thus, we design \AdvTraj to iteratively optimize the attacker's states for each time step that represents the best effort towards the ID-Transfer objective, which also requires less computation resources compared to optimizing over an entire trajectory.\looseness=-1

\vspace{2pt}\noindent\textbf{Adherence to Physical Constraints.} 
The attacker must ensure that the desired input of the bounding box to the association phase can be produced by placing itself at physically realizable positions (\eg not in the sky). This is needed for the attack's practicality and ability to evade anomaly detections. However, enforcing physical constraints on a 2D bounding box sequence is complex, inefficient, and requires extensive knowledge of specific scene topologies. Therefore, instead of optimizing 2D bounding boxes and then translating pixel coordinates into physical locations, we leverage the knowledge of camera parameters\footnote{The camera parameters are specified by the projection matrix, which consists of the extrinsic and intrinsic matrices. The former relates to the camera's location and angle and the latter is affected only by the camera's internal configurations such as the lens. They can be obtained by performing the standard camera calibration process~\cite{cameraCalibration}.} %
to create a differentiable mapping between the physical 3D coordinates\footnote{An attacker can obtain such information by using a drone-mounted stereo camera or LiDAR to estimate the 3D bounding boxes~\cite{muller22TrackerHijacking}.} 
and the 2D bounding boxes perceived by the system. This enables us to \emph{directly optimize over intra-frame physical center displacement} for the attacker to represent the desired movement. Physical constraints can thus be incorporated by clipping the movement to be within a realizable range.\looseness-1

\subsection{Attack Methodology}

We begin by reviewing the MOT association module to identify the conditions required for ID-Transfer. The MOT association module links detected objects (bounding boxes) to existing trackers (predicting current bounding box for associated objects) using a weighted bipartite matching algorithm (\eg Hungarian Algorithm~\cite{hungarian}) that minimizes the \emph{sum of distance metrics} such as Intersection-over-Union (IoU) or its variants~\cite{zheng2020diou} $d:\mathbb{R}^4\times\mathbb{R}^4\to [0,1]$, for all combinations of detected and tracker predicted bounding boxes. The resulting bipartite matching represents assigning unique tracker IDs to detections. Thus, given the set of all existing trackers $\mathcal T\supset\{T_{\IDa},T_{\IDb}\}$ and detected objects $\mathcal O\supset\{O_a,O_b\}$, the following set of conditions implies ID-Transfer between $O_a \text{ and } O_b$ such that $f(O_a)=\IDb\wedge f(O_b)=\IDa$\footnote{For convenience, denote the distance metric between objects $i,j$: $d(i,j)\equiv d(\mathbf{x}_i,\mathbf{x}_j):=d\Big((u_i,v_i,s_i,r_i),(u_j,v_j,s_j,r_j)\Big)$, where $\mathbf{x}=(u,v,s,r,\Vec{u},\Vec{v},\Vec{s})^T$.}:

\begin{equation*}
\begin{aligned}
    &d(\mathbf{x}_a,\hat{\mathbf{x}}_\IDb)+d(\mathbf{x}_b,\hat{\mathbf{x}}_\IDa)<d(\mathbf{x}_a,\hat{\mathbf{x}}_\IDa)+d(\mathbf{x}_b,\hat{\mathbf{x}}_\IDb) & \mathtt{\;(C1)},\\
    &d(\mathbf{x}_b,\hat{\mathbf{x}}_\IDa)<d(\mathbf{x}_i,\hat{\mathbf{x}}_\IDa),\forall i:O_i\in\mathcal O\setminus\{O_a,O_b\} & \mathtt{\;(C2)},\\
    &d(\mathbf{x}_a,\hat{\mathbf{x}}_\IDb)<d(\mathbf{x}_i,\hat{\mathbf{x}}_\IDb),\forall i:O_i\in\mathcal O\setminus\{O_a,O_b\} & \mathtt{\;(C3)}.
\end{aligned}
\end{equation*}

$\mathtt{C1}$ promotes exchanged IDs between $O_a,O_b$, whereas $\mathtt{C2}$ and $\mathtt{C3}$ jointly guarantee that no detected objects other than $O_a,O_b$ will be assigned to trackers $T_a,T_b$. Under our realistic threat model that the attacker can only control its own movement, $\mathtt{C2}$ and $\mathtt{C3}$ can be relaxed in common cases where all tracked entities other than the attacker are benign and have movement patterns consistent with the system's motion prediction model. Thus, to greedily induce ID-Transfer between $O_a \text{ and } O_b$, the attacker needs to achieve $\mathtt{C1}$, which represents a necessary but almost sufficient condition for the objective. Notice that, to cause incorrect ID assignments, it is not sufficient for the attacker to solely minimize $d(\mathbf{x}_a,\hat{\mathbf{x}}_\IDb)$ by placing itself close to the target tracker's predicted states because the matching algorithm optimizes for the \emph{sum} of distance metrics. Therefore, the attacker also needs to maneuver along a trajectory that leads its own tracker predictions close the target object at the same time, \ie minimizing $d(\mathbf{x}_b,\hat{\mathbf{x}}_\IDa)$.

\begin{figure}[t]
\centering
\includegraphics[width=1.0\columnwidth]{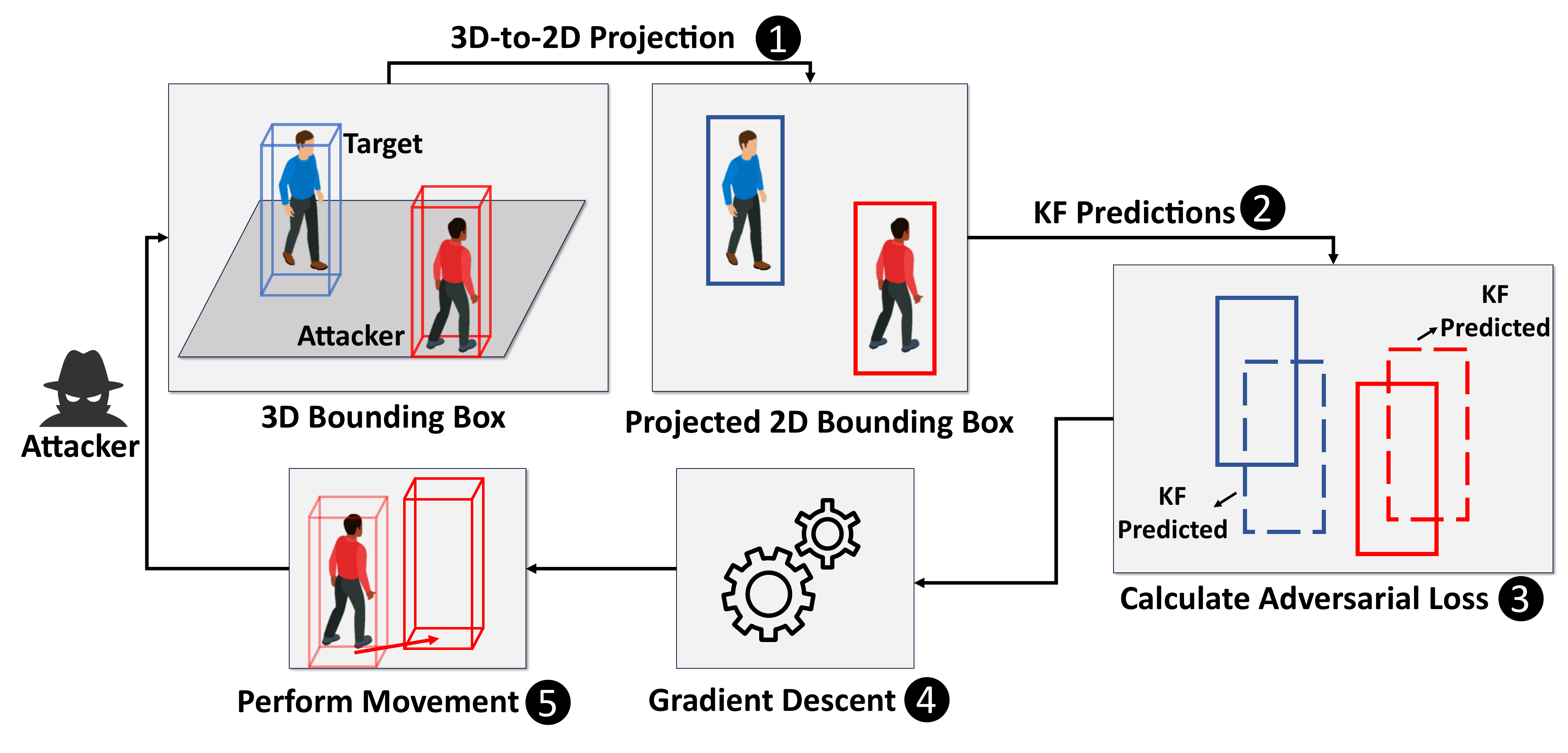}
\caption{Illustration of \AdvTraj's stages.}
\label{fig:attack_pipeline}
\end{figure}

Figure~\ref{fig:attack_pipeline} presents the stages of \AdvTraj, where the attacker iteratively finds a physical location to move towards at each time step through optimizing an adversarial loss function designed to encourage $\mathtt{C1}$ and hence inducing ID-Transfer. Specifically,
the attacker aims to optimize for a real-world location $(x^t_a,y^t_a)$ so that the system perceived states of the attacker minimizes the distance metrics of the transferred ID assignments: $d(\mathbf{x}_a, \hat{\mathbf{x}}_\IDb)+d(\mathbf{x}_v, \hat{\mathbf{x}}_\IDa)$. In other words, \AdvTraj solves:
\begin{equation}
\begin{aligned}
    &\arg\min_{(x^t_a,y^t_a)}d(\mathbf{x}_a, \hat{\mathbf{x}}_\IDb)+d(\mathbf{x}_b, \hat{\mathbf{x}}_\IDa)\\
    &\text{subject to } \|(x^t_a,y^t_a)-(x^{t-1}_a,y^{t-1}_a)\|\leq \epsilon.
\end{aligned}
\end{equation}
The constraint represents the limited physical maneuverability of the attacker such that its movement across frames is bounded within a maximum center displacement value $\epsilon$.

Note that there is a gap between the desired real-world coordinates $(x_a^t,y_a^t)$ representing the attacker's location on the physical ground plane and the detection results $\mathbf{d}=(u_1,v_1,u_2,v_2)$ (the upper-left and bottom-right corners) in the image 2D space fed into the MOT algorithm. Therefore, to directly optimize for physical coordinates, the attacker creates a differentiable mapping between the 3D bounding box vertices in the physical world with its 2D bounding box measurement in the image plane, by performing world-to-image projection of its 3D bounding box vertices (\circled{1}). In other words, given $\mathbf{S}=\{(x_i,y_i,z_i)^T|1\leq i\leq 8\}\in \mathbb{R}^{3\times 8}$, the set of eight corners of the attacker's 3D bounding box, and the camera parameters (\ie projection matrix $\mathbf{P}$), the attacker can obtain its 2D bounding box by a 3D-to-2D projection $\mathcal P:\mathbb{R}^{3\times 8}\rightarrow \mathbb{R}^{4}$ such that:
\begin{equation}\label{projection}
\begin{aligned}
    &\mathbf{d}=\mathcal{P}(\mathbf{S}=(x_i,y_i,z_i)^T|1\leq i\leq 8)\\
    &\quad=\left( \min_{1\leq i\leq 8} x_i'/z_i', \min_{1\leq i\leq 8} y_i'/z_i', \max_{1\leq i\leq 8} x_i'/z_i', \max_{1\leq i\leq 8} y_i'/z_i'\right) \text{\footnotemark}\\
    &\text{where }\begin{pmatrix}
        x_1' & \dots & x_8'\\
        y_1' & \dots & y_8'\\
        z_1' & \dots & z_8'\\
    \end{pmatrix}=\mathbf P
    \begin{pmatrix}
        x_1 & \dots & x_8\\
        y_1 & \dots & y_8\\
        z_1 & \dots & y_8\\
        1   & \dots & 1 
    \end{pmatrix}
    =\mathbf P\begin{pmatrix}
        \mathbf{S} \\ \mathbf{1}
    \end{pmatrix}.
\end{aligned}
\footnotetext{$z_i'$ is the scalar value of 2D homogeneous point representation (with 2 DoF) such that $(x_i'/z_i',y_i'/z_i',1)=(x_i',y_i',z_i')$.}
\end{equation}

For brevity, we denote $\mathbf{\delta}=(\Delta x, \Delta y)$ as the physical center displacement of the attacker along the ground plane representing the applied movement, where $\mathbf{S}+\delta:= \{(x_i+\Delta x, y_i+\Delta y, z_i)|1\leq i\leq 8\}$ is the 3D bounding box after movement, assuming that the physical height of the object stays constant (\circled{5}). Thus far, to achieve ID-Transfer with the target, the attacker finds its desired location to move towards at each time step $t$ by performing gradient descent on the center displacement $\mathbf\delta$ with respect to an adversarial loss (\circled{3}-\circled{4}):\looseness=-1 
\begin{equation}\label{loss}
\begin{aligned}
    &\mathcal{L}(\mathcal P(\mathbf{S}^{t-1}+\mathbf\delta))=d(g(\mathbf{d}^t), \hat{\mathbf{x}}^t_b)+d(\hat{\mathbf{x}}^{t+1}_b,\hat{\mathbf{x}}^{t+1})\\
    &\text{subject to }\|\mathbf{\delta}\|\leq \epsilon 
\end{aligned}
\end{equation}
where $(1)$ $g:\mathbb{R}^{4}\rightarrow\mathbb{R}^4$ is the bijective mapping from the two-corner notation to the center-scale-ratio notation for bounding boxes
, $(2)$ $ \mathbf{d}^t=\mathcal P(\mathbf{S}^{t-1}+\mathbf\delta)$ is the attacker's projected 2D bounding box corresponding to physical movement $\delta=(\Delta x, \Delta y)$, $(3)$ $\hat{\mathbf{x}}^{t+1}_b=\mathbf{F}\hat{\mathbf{x}}_b^t$ is the target's prior 
state estimates for 
$t+1$, and $(4)$ $\hat{\mathbf{x}}^{t+1}=\mathbf{F} [\mathtt{KF}_{\mathtt{ID}_a}.\mathtt{update}(\mathbf{x}^t)]$ is the product between the transition matrix and the posterior estimation after update (\ie the attacker's KF prior predicted states if updated by state observation $g^{-1}(\mathbf{x}^t)$). Note that this single-step KF prediction is linear~\cite{kalman1960new} hence differentiable (\circled{2}).\looseness=-1

With the adversarial loss function defined, the attacker performs gradient descent (using readily available optimizers like Adam~\cite{adam}) on $\mathbf\delta$, which represents the desired movement on the ground plane, clips the result to be within physically realizable regions, and maneuvers towards the target waypoint. The series of waypoints form an adversarial trajectory that encourages ID-Transfer. The complete \AdvTraj attack process is summarized in Algorithm~\ref{alg:id_swap} and the intuitive explanation of the loss function is presented in Appendix~\ref{intuition}.

\begin{figure}[t]
\begin{algorithm}[H] 
\setstretch{1}
    \small
	\caption{\AdvTraj ID-Transfer Trajectory Generator}
	\label{alg:id_swap}
	\begin{algorithmic}[1]
        \Require{Attacker 3D bounding box $\mathbf{S}^{t-1}$, KF trackers $\mathtt{KF_{ID_b}}, \mathtt{KF_{ID_a}}$, maximum displacement $\epsilon$, number of iterations $\mathtt{iter}$}
		\Ensure{Attacker center displacement $\delta$}
		\vspace{0.3mm}
		\Function{AdvTraj-ID-Transfer}{$\mathbf{S}^{t-1}, \mathtt{KF_{ID_b}}, \mathtt{KF_{ID_a}}$}
        \State Initialize $\delta = (0,0)$
        \State $\hat{\mathbf{x}}_b^t=\mathtt{KF_{ID_b}.predict()}$
        \State $\hat{\mathbf{x}}_a^t=\mathtt{KF_{ID_a}.predict()}$
        \If{$d(a,\mathtt{ID_b})+d(b,\mathtt{ID_a})>d(a,\mathtt{ID_a})+d(b, \mathtt{ID_b})$}
            \State $\hat{\mathbf{x}}^{t+1}_b=\mathbf{F}\hat{\mathbf{x}}_b^t$
            \State Initialize $\mathtt{i}=0$
            \While{$\mathtt{i< iter}$}
                \State $\Delta=\nabla_{\mathbf{\delta}}\mathcal{L}(\mathcal{P}(\mathbf{S}^{t-1}+\mathbf{\delta}))$
                \State $\mathbf{\delta} = \mathtt{GradientDescent}(\mathbf{\delta}, \Delta)$
                \State $\mathtt{i}=\mathtt{i}+1$
            \EndWhile
            \State $\mathbf{\delta}=\mathtt{Clip}(\mathbf{\delta}, \epsilon)$
        \EndIf
        \State \Return{$\mathbf{\delta}$}
        \EndFunction
         \While{IDs are not transferred and a new frame arrives}
     \State $\mathbf{\delta} = \mathtt{AdvTraj-ID-Transfer}(\mathbf{S}^{t-1}, \mathtt{KF_{ID_b}}, \mathtt{KF_{ID_a}})$
     \State $\mathtt{PerformMovement}(\mathbf{\delta})$
     \EndWhile
	\end{algorithmic}
\end{algorithm}
\end{figure}

\subsection{Universal Adversarial Maneuvers}\label{black-box-attack}

Although \AdvTraj automatically generates adversarial trajectories online, it may be challenging for non-automated agents (\eg humans) to calculate and physically follow the crafted trajectory in real-time, since it would require fine-grained motor control. Thus, based on the understanding of the vulnerability in MOT algorithms that are exploited by \AdvTraj, we propose universal adversarial maneuvers (UAMs) that are practically executable by human walkers/drivers.\looseness=-1

To achieve this, we start by abstracting the adversarial trajectory generation process, and then investigate the patterns of the generated trajectories (in the 2D plane) to find real-world applicable maneuvers. Although \AdvTraj functions regardless of the target's movement, to extract UAMs for common scenarios, we consider a target that moves at a constant velocity (speed and direction), which is the dominant pattern for benign moving objects in daily scenarios. We randomize the starting positions of the attacker relative to the target and perform the attack as the target moves. Detailed analysis and generated trajectories are shown in Appendix~\ref{uni_adv_traj}.\looseness=-1

We observe that the generated trajectories share a prominent pattern where the attacker initially attempts to close its distance to the target greedily regardless of the initial position ($\mathtt{P1}$), and performs maneuvers along the same direction as the target with varying speed (non-linear movement) after being around the target's location ($\mathtt{P2}$). Furthermore, these adversarial trajectories can be divided into two categories, where the attacker reaches the target's history path (behind the target) or the target's projected path (ahead of the target), before the ID-Transfer. By isolating these two common cases that other adversarial trajectories can reduce to, we further examine the attacker's specific movement patterns, which can be summarized as follows.\looseness=-1 

\begin{figure*}[t]
\centering
\begin{tabular}[t]{c|c}
 \begin{tabular}{c}
        \begin{subfigure}[t]{0.4\textwidth}
            \centering
            \includegraphics[width=1.0\linewidth]{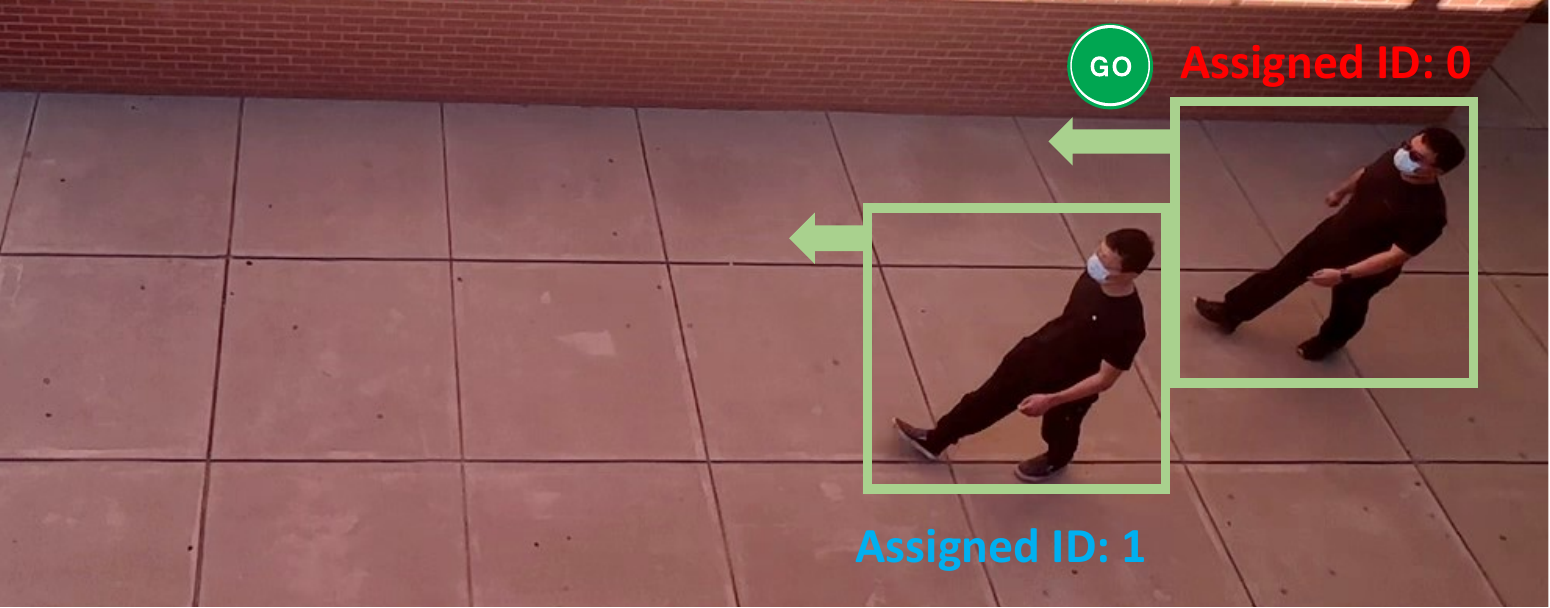}
        \end{subfigure}\\
        \begin{subfigure}[t]{0.4\textwidth}
            \centering
            \includegraphics[width=1.0\linewidth]{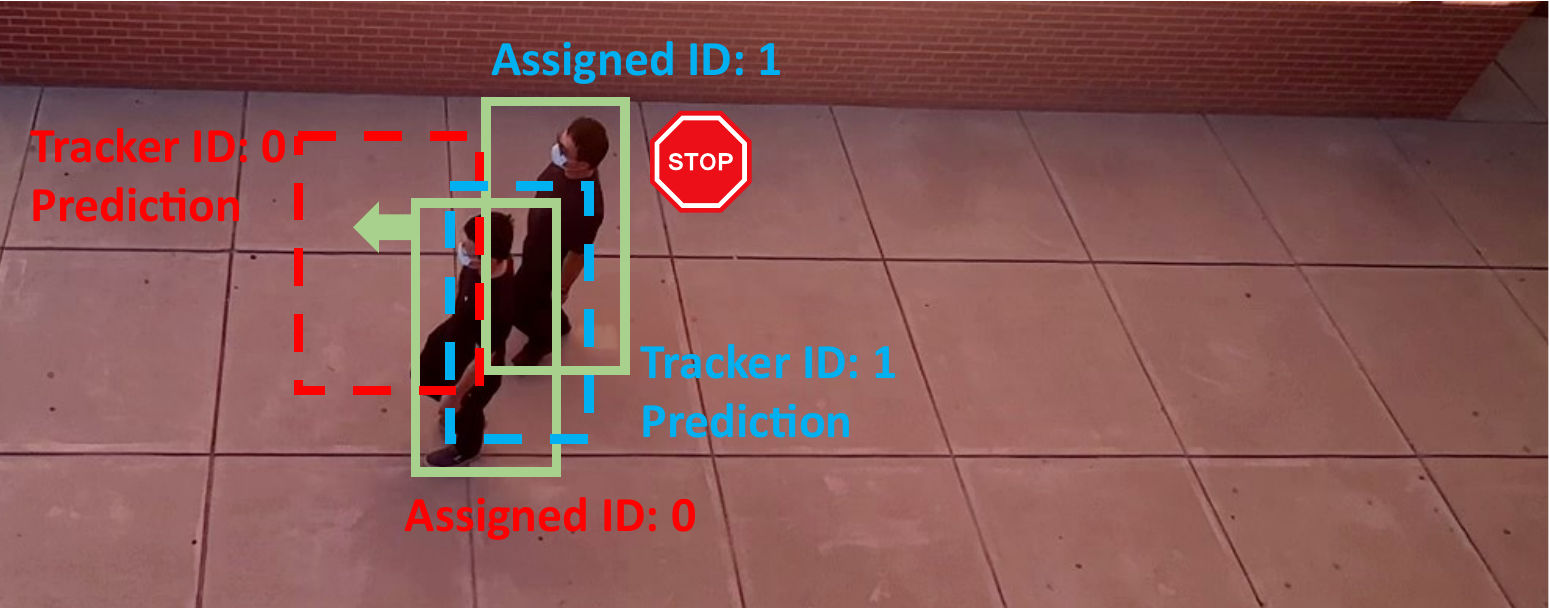}
            \caption{Attacker starts behind the target (Go-and-Stop)}
        \end{subfigure}\\
        \end{tabular}
     &  
        \begin{tabular}{c}
        \begin{subfigure}[t]{0.4\textwidth}
            \centering
            \includegraphics[width=1.0\linewidth]{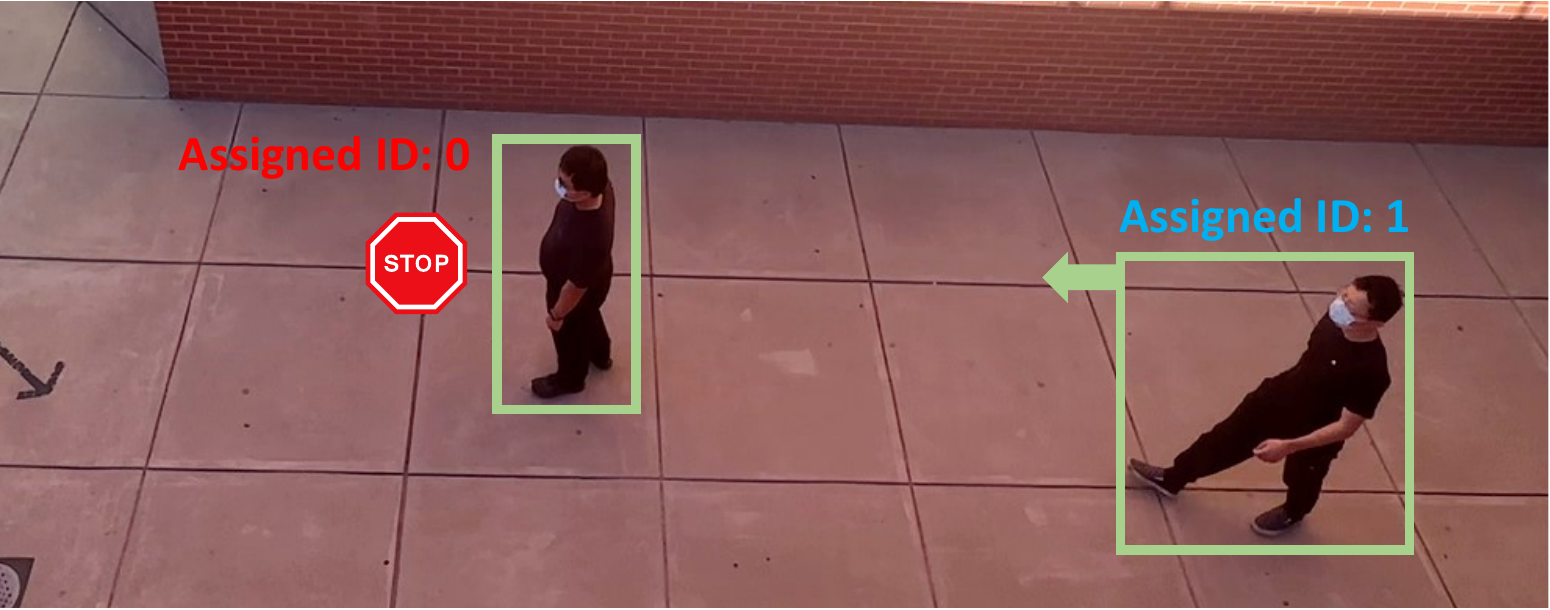}
        \end{subfigure}\\
        \begin{subfigure}[t]{0.4\textwidth}
            \centering
            \includegraphics[width=1.0\linewidth]{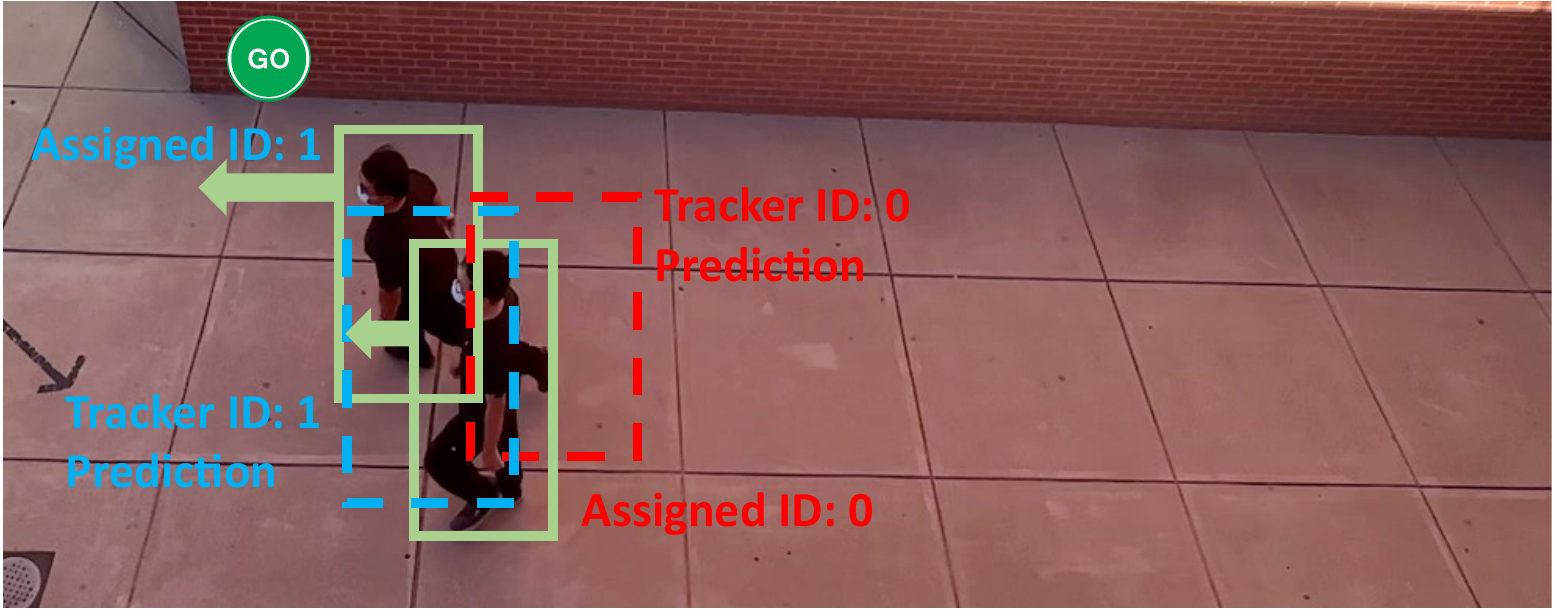}
            \caption{Attacker starts ahead of the target (Stop-and-Go)}
        \end{subfigure}\\
        \end{tabular}\\ 
\end{tabular}
\caption{Illustration of the two universal adversarial maneuvers for ID-Transfer. Top/lower walker is the attacker/non-cooperating target, respectively. Solid green boxes represent the detection results of the OD model, while dashed bounding boxes represent predictions made by the KF motion prediction module of the MOT system.} 
\label{fig:universal}
\end{figure*}

When the attacker starts behind the target, it approaches the target at a higher speed and converges to the same moving direction as the target ($\mathtt{P1}$).  As the attacker approaches the target, it abruptly decelerates, positioning itself behind the target ($\mathtt{P2}$). However, the KF prediction, based on the attacker's previous high-speed trajectory, would predict the attacker to be ahead of itself but close to the target. Therefore, the sum of distance metrics becomes lower for the attacker being assigned to the target's tracker and vice versa.\looseness=-1

For the attacker initiated ahead of the target, it moves towards the target's projected path and moves in the exact opposite direction as the target (due to greedy reduction in distance) ($\mathtt{P1}$). When the target passes by, the attacker turns to move in the same direction as the target but at a higher speed when the target passes ($\mathtt{P2}$). This sudden acceleration also places the attacker's actual position close to the target's KF-predicted location, while its own KF prediction (lagging behind) is close to the target's actual position. This also results in the sum of distance metrics becoming lower for the attacker being assigned to the target's tracker and vice versa.\looseness=-1

These two patterns of adversarial trajectories can be summarized into two UAMs, namely \emph{Go-and-Stop} and \emph{Stop-and-Go}, which are visually demonstrated in Figure~\ref{fig:universal}. The attacker starts by closing its distance from the target by moving towards the target's history path (behind) or projected path (ahead), then employs the corresponding tactic to complete the ID-Transfer attack. Both UAMs increase $d(a,\IDa)$ drastically by inducing a large difference between $\mathbf{d}_a$ and $\hat{\mathbf{d}}_a$ when $\hat{\mathbf{d}}_a$ is expected to be close to the actual states of the target object $\mathbf{d}_b$. Since weighted bipartite matching aims to minimize the global assignment cost, misalignment $d(a,\IDa)+d(b,\IDb)>d(a,\IDb)+d(b,\IDa)$ in predicted and actual positions can deceive the system into incorrectly associating the identities.


\section{Evaluation}\label{sec:eval}

We evaluate \AdvTraj by simulating the $(\mathtt{T1})$ optimized adversary (\eg automated agents) in the open source simulator CARLA~\cite{Dosovitskiy17} and conducting real-world experiments for the $(\mathtt{T2})$ heuristic adversary (\eg human walkers/drivers) under various application scenarios of MOT systems. 

\subsection{Evaluation Methodology and Setup}

\vspace{2pt}\noindent\textbf{Evaluation Metrics.} We measure the effectiveness of \AdvTraj by the \emph{attack success rate} (ASR) as the number of successful attacks over the total number of randomized simulations/real-world experiments. For the $(\mathtt{T1})$ optimized adversaries, an attack is considered successful if the IDs of the attacker and the target object assigned by the MOT system are swapped and stay exchanged after the attack, as compared to the initial assignment when their trackers are created.
For the $(\mathtt{T2})$ heuristic adversaries, the ASR is calculated as the number of successful attacks over the total number of attacks performed in the whole continuous video recorded for each scenario where the attacker does not have the output from the system when performing the universal adversarial maneuvers (UAMs). Since naturally occurring missed detections during non-line-of-sight occlusions may lead to tracker loss, we consider an attack to be successful if the ID of the attacker is transferred to the target or the target's ID is transferred to the attacker after the attack. This is a stronger definition than the ID-Switch attack used in previous works on attacking MOT ~\cite{HijackingTracker, Lin2021TraSw, Ma2023Attack, FFAttack} since we require the preservation of at least one original ID of the attacker or target which is assigned incorrectly due to the attack.\looseness=-1

\vspace{2pt}\noindent\textbf{MOT Algorithms.} Although different MOT algorithms vary in specific designs (e.g., association distance metrics, estimated states, KF parameters), most state-of-the-art models adopt the tracking-by-detection framework and use KF (or its variant) as the motion prediction module, which can be attacked using \AdvTraj by employing the corresponding distance metrics and KF parameters in Equation~\ref{loss}. However, these common design principles also allow the black-box attacker to perform transfer attacks using surrogate models. Thus, to evaluate the effectiveness of white-box attacks (by $(\mathtt{T1})$ adversary), we implement \AdvTraj against the representative tracking-by-detection algorithm~SORT\cite{SORT} by writing the attack module, simulation APIs, the SORT algorithm and its KF dependencies using Python 3.8 and Tensorflow 2.6.0 
totaling 2,585 LoC\footnote{Available at \url{https://github.com/ch3ny1/AdvTraj_ID_Transfer}.}. 

We evaluate the transferability of the adversarial trajectories generated against SORT to other MOT algorithms to assess the effectiveness of black-box attacks by $(\mathtt{T1})$ adversary. Specifically, we consider five other state-of-the-art MOT algorithms (detailed in Table~\ref{tab:mot_table}): 
OC-SORT~\cite{OCSORT}, ByteTrack~\cite{ByteTrack}, BoT-SORT ~\cite{BoTSORT}, and StrongSORT ~\cite{strongsort}. We use the respective default parameters of each algorithm for evaluation. For trackers that require appearance descriptors, we use the same pre-trained OSNET~\cite{ICCV19OSNet} as the ReID backbone for fair comparison.\looseness=-1

On the other hand, the $(\mathtt{T2})$ adversary performing UAMs do not have the capability to conduct real-time optimization and fine-grained motor control. Thus, we conduct real-world experiments for black-box ID-Transfer attacks by $(\mathtt{T2})$ adversary and evaluate the recorded footage on each of the six MOT algorithms.

\begin{table}[t]
\caption{Details of evaluated MOT algorithms.}\label{tab:mot_table}
\centering
\resizebox{\columnwidth}{!}{
\begin{threeparttable}
\renewcommand{\arraystretch}{1.2}
\setlength{\tabcolsep}{4pt}
\begin{tabular}{l|c|c|c|}
\cline{2-4}
& \textbf{Motion Prediction Model}  & \textbf{ReID Included} & \textbf{IDF1$\uparrow$}\tnote{\textdagger} \\ \hline
\multicolumn{1}{|l|}{\textbf{SORT}~\cite{SORT}} & \multirow{5}{*}{Std. Kalman Filter} & N & 76.9 \\ \cline{1-1}\cline{3-4}
\multicolumn{1}{|l|}{\textbf{OC-SORT}~\cite{OCSORT}} &  & N & 77.5 \\ \cline{1-1}\cline{3-4}
\multicolumn{1}{|l|}{\textbf{ByteTrack}~\cite{ByteTrack}} &  & N\tnote{\textdagger\textdagger} & 79.3 \\ \cline{1-1}\cline{3-4}
\multicolumn{1}{|l|}{\textbf{Deep OC-SORT}~\cite{DeepOCSORT}} &  & Y & 80.6 \\ \cline{1-1}\cline{3-4}
\multicolumn{1}{|l|}{\textbf{BoT SORT}~\cite{BoTSORT}} &  & Y & 80.2 \\ \hline
\multicolumn{1}{|l|}{\textbf{StrongSORT}~\cite{strongsort}} & NSA Kalman Filter & Y & 82.3 \\ \hline
\end{tabular}
\begin{tablenotes}[flushleft]
    \item[\textdagger] Ratio of correctly identified detections over the average of ground truth and predicted detections, evaluated on MOT17~\cite{MOT17}.
    \item[\textdagger\textdagger] ReID can be incorporated, though it is shown that the best performing model uses IoU only~\cite{ByteTrack}, which is also the original implementation by the authors.
\end{tablenotes}
\end{threeparttable}
}
\end{table}

\vspace{2pt}\noindent\textbf{Scenario Setup.} We evaluate \AdvTraj in two
applications: pedestrian/vehicle surveillance and autonomous driving (AD). These scenarios were chosen to represent the safety-critical applications of MOT systems with the consequences of losing track of a target of interest and unsafe driving decisions when the attack occurs. We primarily focus on the pedestrian surveillance and AD scenarios in this section, where we employ walkers as tracked objects\footnote{Agents in CARLA simulator are available as either walkers or vehicles.}. The setup and results on vehicle surveillance are presented in Appendix~\ref{vehicle_surveillance}.\looseness=-1

For the pedestrian surveillance scenario, a fixed camera is mounted at a high position (e.g., edge of a building) as the input sensor to the MOT algorithm. It provides wide views with less occlusion of the open walking area beneath it.\looseness=-1

For the AD scenario, the MOT system is deployed as part of the perception module of the AD system, where the image sensor captures video streams as input. We consider two representative patterns for vehicle-pedestrian interactions: ($1$) pedestrians walk \emph{ perpendicular} to a stopped AD, which usually occurs at a crosswalk; ($2$) pedestrians on the sidewalk walking \emph{parallel} to the AD, which drives forward at a constant speed of 20 KM/h.\looseness=-1

\begin{figure*}[t]
\begin{subfigure}{.33\textwidth}
  \centering
\begin{tikzpicture}

\definecolor{crimson2143940}{RGB}{214,39,40}
\definecolor{darkgray176}{RGB}{176,176,176}
\definecolor{darkorange25512714}{RGB}{255,127,14}
\definecolor{forestgreen4416044}{RGB}{44,160,44}
\definecolor{lightgray204}{RGB}{204,204,204}
\definecolor{mediumpurple148103189}{RGB}{148,103,189}
\definecolor{steelblue31119180}{RGB}{31,119,180}

\begin{axis}[
width=6cm,
height=5cm,
legend cell align={left},
legend style={
  fill opacity=0.8,
  draw opacity=1,
  text opacity=1,
  at={(0.97,0.03)},
  anchor=south east,
  draw=lightgray204,
  nodes={scale=0.7, transform shape}
},
tick align=outside,
tick pos=left,
x grid style={darkgray176},
xmin=-0.3625, xmax=5.9625,
xtick style={color=black},
xtick={0.15,1.15,2.15,3.15,4.15,5.15},
xticklabel style={rotate=45.0},
xticklabels={\footnotesize SORT, \footnotesize ByteTrack, \footnotesize OC-SORT, \footnotesize Deep OC-SORT, \footnotesize BoT-SORT, \footnotesize StrongSORT},
y grid style={darkgray176},
ylabel={\footnotesize (Transfer) Attack Success Rate (\%)},
ymin=0, ymax=105,
ytick style={color=black}
]
\draw[draw=none,fill=steelblue31119180] (axis cs:-0.075,0) rectangle (axis cs:0.075,100);
\addlegendimage{ybar,ybar legend,draw=none,fill=steelblue31119180}
\addlegendentry{0.5 m/s}

\draw[draw=none,fill=steelblue31119180] (axis cs:0.925,0) rectangle (axis cs:1.075,74);
\draw[draw=none,fill=steelblue31119180] (axis cs:1.925,0) rectangle (axis cs:2.075,73);
\draw[draw=none,fill=steelblue31119180] (axis cs:2.925,0) rectangle (axis cs:3.075,91);
\draw[draw=none,fill=steelblue31119180] (axis cs:3.925,0) rectangle (axis cs:4.075,93);
\draw[draw=none,fill=steelblue31119180] (axis cs:4.925,0) rectangle (axis cs:5.075,91);
\draw[draw=none,fill=darkorange25512714] (axis cs:0.075,0) rectangle (axis cs:0.225,100);
\addlegendimage{ybar,ybar legend,draw=none,fill=darkorange25512714}
\addlegendentry{1.0 m/s}

\draw[draw=none,fill=darkorange25512714] (axis cs:1.075,0) rectangle (axis cs:1.225,57);
\draw[draw=none,fill=darkorange25512714] (axis cs:2.075,0) rectangle (axis cs:2.225,66);
\draw[draw=none,fill=darkorange25512714] (axis cs:3.075,0) rectangle (axis cs:3.225,80);
\draw[draw=none,fill=darkorange25512714] (axis cs:4.075,0) rectangle (axis cs:4.225,78);
\draw[draw=none,fill=darkorange25512714] (axis cs:5.075,0) rectangle (axis cs:5.225,88);
\draw[draw=none,fill=forestgreen4416044] (axis cs:0.225,0) rectangle (axis cs:0.375,100);
\addlegendimage{ybar,ybar legend,draw=none,fill=forestgreen4416044}
\addlegendentry{1.5 m/s}

\draw[draw=none,fill=forestgreen4416044] (axis cs:1.225,0) rectangle (axis cs:1.375,65);
\draw[draw=none,fill=forestgreen4416044] (axis cs:2.225,0) rectangle (axis cs:2.375,72);
\draw[draw=none,fill=forestgreen4416044] (axis cs:3.225,0) rectangle (axis cs:3.375,80);
\draw[draw=none,fill=forestgreen4416044] (axis cs:4.225,0) rectangle (axis cs:4.375,79);
\draw[draw=none,fill=forestgreen4416044] (axis cs:5.225,0) rectangle (axis cs:5.375,89);
\draw[draw=none,fill=crimson2143940] (axis cs:0.375,0) rectangle (axis cs:0.525,100);
\addlegendimage{ybar,ybar legend,draw=none,fill=crimson2143940}
\addlegendentry{2.0 m/s}

\draw[draw=none,fill=crimson2143940] (axis cs:1.375,0) rectangle (axis cs:1.525,66);
\draw[draw=none,fill=crimson2143940] (axis cs:2.375,0) rectangle (axis cs:2.525,67);
\draw[draw=none,fill=crimson2143940] (axis cs:3.375,0) rectangle (axis cs:3.525,83);
\draw[draw=none,fill=crimson2143940] (axis cs:4.375,0) rectangle (axis cs:4.525,80);
\draw[draw=none,fill=crimson2143940] (axis cs:5.375,0) rectangle (axis cs:5.525,92);
\draw[draw=none,fill=mediumpurple148103189] (axis cs:0.525,0) rectangle (axis cs:0.675,100);
\addlegendimage{ybar,ybar legend,draw=none,fill=mediumpurple148103189}
\addlegendentry{2.5 m/s}

\draw[draw=none,fill=mediumpurple148103189] (axis cs:1.525,0) rectangle (axis cs:1.675,61);
\draw[draw=none,fill=mediumpurple148103189] (axis cs:2.525,0) rectangle (axis cs:2.675,68);
\draw[draw=none,fill=mediumpurple148103189] (axis cs:3.525,0) rectangle (axis cs:3.675,68);
\draw[draw=none,fill=mediumpurple148103189] (axis cs:4.525,0) rectangle (axis cs:4.675,67);
\draw[draw=none,fill=mediumpurple148103189] (axis cs:5.525,0) rectangle (axis cs:5.675,65);
\end{axis}

\end{tikzpicture}
\vspace{-15pt}
  \caption{Pedestrian Surveillance}
\end{subfigure}%
\begin{subfigure}{0.33\textwidth}
  \centering
\begin{tikzpicture}

\definecolor{crimson2143940}{RGB}{214,39,40}
\definecolor{darkgray176}{RGB}{176,176,176}
\definecolor{darkorange25512714}{RGB}{255,127,14}
\definecolor{forestgreen4416044}{RGB}{44,160,44}
\definecolor{lightgray204}{RGB}{204,204,204}
\definecolor{mediumpurple148103189}{RGB}{148,103,189}
\definecolor{steelblue31119180}{RGB}{31,119,180}

\begin{axis}[
width=6cm,
height=5cm,
legend cell align={left},
legend style={
  fill opacity=0.8,
  draw opacity=1,
  text opacity=1,
  at={(0.97,0.03)},
  anchor=south east,
  draw=lightgray204
},
tick align=outside,
tick pos=left,
x grid style={darkgray176},
xmin=-0.3625, xmax=5.9625,
xtick style={color=black},
xtick={0.15,1.15,2.15,3.15,4.15,5.15},
xticklabel style={rotate=45.0},
xticklabels={\footnotesize SORT, \footnotesize ByteTrack, \footnotesize OC-SORT, \footnotesize Deep OC-SORT, \footnotesize BoT-SORT, \footnotesize StrongSORT},
y grid style={darkgray176},
ylabel={\footnotesize (Transfer) Attack Success Rate (\%)},
ymin=0, ymax=105,
ytick style={color=black}
]
\draw[draw=none,fill=steelblue31119180] (axis cs:-0.075,0) rectangle (axis cs:0.075,100);
\addlegendimage{ybar,ybar legend,draw=none,fill=steelblue31119180, postaction={         pattern=north east lines     }}

\draw[draw=none,fill=steelblue31119180] (axis cs:0.925,0) rectangle (axis cs:1.075,47);
\draw[draw=none,fill=steelblue31119180] (axis cs:1.925,0) rectangle (axis cs:2.075,85);
\draw[draw=none,fill=steelblue31119180] (axis cs:2.925,0) rectangle (axis cs:3.075,85);
\draw[draw=none,fill=steelblue31119180] (axis cs:3.925,0) rectangle (axis cs:4.075,86);
\draw[draw=none,fill=steelblue31119180] (axis cs:4.925,0) rectangle (axis cs:5.075,82);
\draw[draw=none,fill=darkorange25512714] (axis cs:0.075,0) rectangle (axis cs:0.225,100);
\addlegendimage{ybar,ybar legend,draw=none,fill=darkorange25512714}

\draw[draw=none,fill=darkorange25512714] (axis cs:1.075,0) rectangle (axis cs:1.225,51);
\draw[draw=none,fill=darkorange25512714] (axis cs:2.075,0) rectangle (axis cs:2.225,81);
\draw[draw=none,fill=darkorange25512714] (axis cs:3.075,0) rectangle (axis cs:3.225,88);
\draw[draw=none,fill=darkorange25512714] (axis cs:4.075,0) rectangle (axis cs:4.225,75);
\draw[draw=none,fill=darkorange25512714] (axis cs:5.075,0) rectangle (axis cs:5.225,78);
\draw[draw=none,fill=forestgreen4416044] (axis cs:0.225,0) rectangle (axis cs:0.375,100);
\addlegendimage{ybar,ybar legend,draw=none,fill=forestgreen4416044}

\draw[draw=none,fill=forestgreen4416044] (axis cs:1.225,0) rectangle (axis cs:1.375,38);
\draw[draw=none,fill=forestgreen4416044] (axis cs:2.225,0) rectangle (axis cs:2.375,68);
\draw[draw=none,fill=forestgreen4416044] (axis cs:3.225,0) rectangle (axis cs:3.375,69);
\draw[draw=none,fill=forestgreen4416044] (axis cs:4.225,0) rectangle (axis cs:4.375,95);
\draw[draw=none,fill=forestgreen4416044] (axis cs:5.225,0) rectangle (axis cs:5.375,92);
\draw[draw=none,fill=crimson2143940] (axis cs:0.375,0) rectangle (axis cs:0.525,100);
\addlegendimage{ybar,ybar legend,draw=none,fill=crimson2143940}

\draw[draw=none,fill=crimson2143940] (axis cs:1.375,0) rectangle (axis cs:1.525,43);
\draw[draw=none,fill=crimson2143940] (axis cs:2.375,0) rectangle (axis cs:2.525,49);
\draw[draw=none,fill=crimson2143940] (axis cs:3.375,0) rectangle (axis cs:3.525,60);
\draw[draw=none,fill=crimson2143940] (axis cs:4.375,0) rectangle (axis cs:4.525,67);
\draw[draw=none,fill=crimson2143940] (axis cs:5.375,0) rectangle (axis cs:5.525,69);
\draw[draw=none,fill=mediumpurple148103189] (axis cs:0.525,0) rectangle (axis cs:0.675,95);
\addlegendimage{ybar,ybar legend,draw=none,fill=mediumpurple148103189}

\draw[draw=none,fill=mediumpurple148103189] (axis cs:1.525,0) rectangle (axis cs:1.675,40);
\draw[draw=none,fill=mediumpurple148103189] (axis cs:2.525,0) rectangle (axis cs:2.675,62);
\draw[draw=none,fill=mediumpurple148103189] (axis cs:3.525,0) rectangle (axis cs:3.675,68);
\draw[draw=none,fill=mediumpurple148103189] (axis cs:4.525,0) rectangle (axis cs:4.675,67);
\draw[draw=none,fill=mediumpurple148103189] (axis cs:5.525,0) rectangle (axis cs:5.675,65);
\end{axis}

\end{tikzpicture}
  \vspace{-15pt}
  \caption{AD-Perpendicular}
\end{subfigure}
\begin{subfigure}{0.33\textwidth}
  \centering
\begin{tikzpicture}

\definecolor{crimson2143940}{RGB}{214,39,40}
\definecolor{darkgray176}{RGB}{176,176,176}
\definecolor{darkorange25512714}{RGB}{255,127,14}
\definecolor{forestgreen4416044}{RGB}{44,160,44}
\definecolor{lightgray204}{RGB}{204,204,204}
\definecolor{mediumpurple148103189}{RGB}{148,103,189}
\definecolor{steelblue31119180}{RGB}{31,119,180}

\begin{axis}[
width=6cm,
height=5cm,
legend cell align={left},
legend style={
  fill opacity=0.8,
  draw opacity=1,
  text opacity=1,
  at={(0.97,0.03)},
  anchor=south east,
  draw=lightgray204
},
tick align=outside,
tick pos=left,
x grid style={darkgray176},
xmin=-0.3625, xmax=5.9625,
xtick style={color=black},
xtick={0.15,1.15,2.15,3.15,4.15,5.15},
xticklabel style={rotate=45.0},
xticklabels={\footnotesize SORT, \footnotesize ByteTrack, \footnotesize OC-SORT, \footnotesize Deep OC-SORT, \footnotesize BoT-SORT, \footnotesize StrongSORT},
y grid style={darkgray176},
ylabel={\footnotesize (Transfer) Attack Success Rate (\%)},
ymin=0, ymax=105,
ytick style={color=black}
]
\draw[draw=none,fill=steelblue31119180] (axis cs:-0.075,0) rectangle (axis cs:0.075,100);
\addlegendimage{ybar,ybar legend,draw=none,fill=steelblue31119180}

\draw[draw=none,fill=steelblue31119180] (axis cs:0.925,0) rectangle (axis cs:1.075,74);
\draw[draw=none,fill=steelblue31119180] (axis cs:1.925,0) rectangle (axis cs:2.075,73);
\draw[draw=none,fill=steelblue31119180] (axis cs:2.925,0) rectangle (axis cs:3.075,91);
\draw[draw=none,fill=steelblue31119180] (axis cs:3.925,0) rectangle (axis cs:4.075,93);
\draw[draw=none,fill=steelblue31119180] (axis cs:4.925,0) rectangle (axis cs:5.075,91);
\draw[draw=none,fill=darkorange25512714] (axis cs:0.075,0) rectangle (axis cs:0.225,100);
\addlegendimage{ybar,ybar legend,draw=none,fill=darkorange25512714}

\draw[draw=none,fill=darkorange25512714] (axis cs:1.075,0) rectangle (axis cs:1.225,57);
\draw[draw=none,fill=darkorange25512714] (axis cs:2.075,0) rectangle (axis cs:2.225,66);
\draw[draw=none,fill=darkorange25512714] (axis cs:3.075,0) rectangle (axis cs:3.225,80);
\draw[draw=none,fill=darkorange25512714] (axis cs:4.075,0) rectangle (axis cs:4.225,78);
\draw[draw=none,fill=darkorange25512714] (axis cs:5.075,0) rectangle (axis cs:5.225,88);
\draw[draw=none,fill=forestgreen4416044] (axis cs:0.225,0) rectangle (axis cs:0.375,100);
\addlegendimage{ybar,ybar legend,draw=none,fill=forestgreen4416044}

\draw[draw=none,fill=forestgreen4416044] (axis cs:1.225,0) rectangle (axis cs:1.375,65);
\draw[draw=none,fill=forestgreen4416044] (axis cs:2.225,0) rectangle (axis cs:2.375,72);
\draw[draw=none,fill=forestgreen4416044] (axis cs:3.225,0) rectangle (axis cs:3.375,80);
\draw[draw=none,fill=forestgreen4416044] (axis cs:4.225,0) rectangle (axis cs:4.375,79);
\draw[draw=none,fill=forestgreen4416044] (axis cs:5.225,0) rectangle (axis cs:5.375,89);
\draw[draw=none,fill=crimson2143940] (axis cs:0.375,0) rectangle (axis cs:0.525,100);
\addlegendimage{ybar,ybar legend,draw=none,fill=crimson2143940}

\draw[draw=none,fill=crimson2143940] (axis cs:1.375,0) rectangle (axis cs:1.525,66);
\draw[draw=none,fill=crimson2143940] (axis cs:2.375,0) rectangle (axis cs:2.525,67);
\draw[draw=none,fill=crimson2143940] (axis cs:3.375,0) rectangle (axis cs:3.525,83);
\draw[draw=none,fill=crimson2143940] (axis cs:4.375,0) rectangle (axis cs:4.525,80);
\draw[draw=none,fill=crimson2143940] (axis cs:5.375,0) rectangle (axis cs:5.525,92);
\draw[draw=none,fill=mediumpurple148103189] (axis cs:0.525,0) rectangle (axis cs:0.675,100);
\addlegendimage{ybar,ybar legend,draw=none,fill=mediumpurple148103189}

\draw[draw=none,fill=mediumpurple148103189] (axis cs:1.525,0) rectangle (axis cs:1.675,61);
\draw[draw=none,fill=mediumpurple148103189] (axis cs:2.525,0) rectangle (axis cs:2.675,68);
\draw[draw=none,fill=mediumpurple148103189] (axis cs:3.525,0) rectangle (axis cs:3.675,68);
\draw[draw=none,fill=mediumpurple148103189] (axis cs:4.525,0) rectangle (axis cs:4.675,67);
\draw[draw=none,fill=mediumpurple148103189] (axis cs:5.525,0) rectangle (axis cs:5.675,65);
\end{axis}

\end{tikzpicture}
  \vspace{-15pt}
  \caption{AD-Parallel}
\end{subfigure}
\caption{White-box attack success rate on SORT and black-box transfer attack success rates on other MOT algorithms of \AdvTraj generated adversarial trajectories with respect to different target movement speeds.}
\label{fig:diff_speed}
\end{figure*}
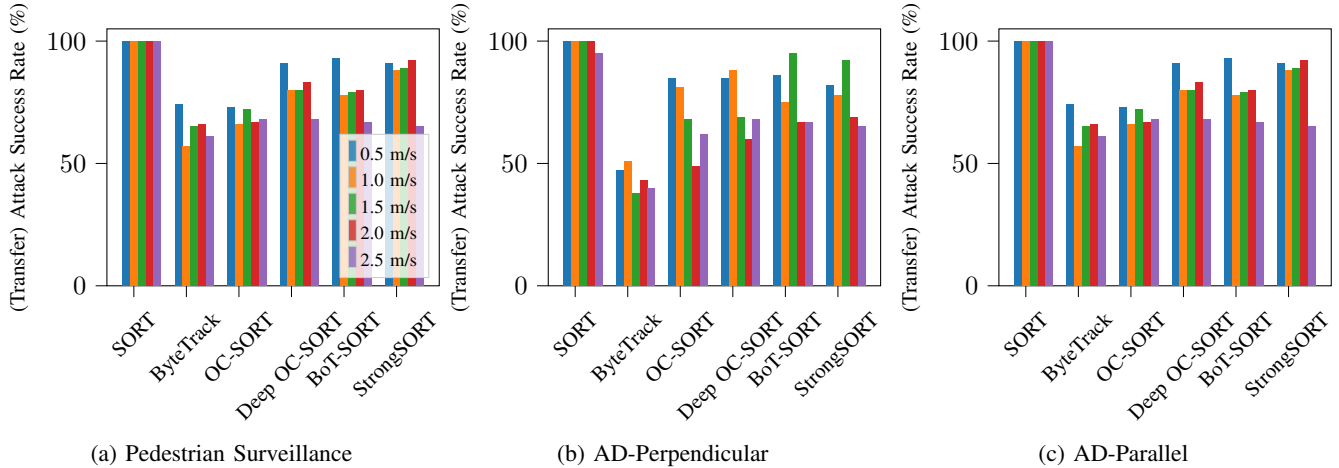

\vspace{2pt}\noindent\textbf{Simulation Setup.} To simulate the $(\mathtt{T1})$ adversary and evaluate the effectiveness of \AdvTraj, we leverage the CARLA simulator, known for its physics-compliant engine and photorealistic rendering. The camera parameters, captured video frames, 3D coordinates, and bounding boxes of detected objects can be extracted directly from CARLA. We choose appropriate locations in CARLA Town 01/05 and perform 100 randomized simulations for each of the three scenarios: ($1$) pedestrian surveillance, ($2$) AD-parallel, and ($3$) AD-perpendicular. The MOT system retrieves $1920\times1080$ RGB images and object bounding boxes at 10 frames per second from CARLA and performs real-time ID assignments. For each simulation, the spawn locations for the attacker and target are randomly chosen in walkable/drivable areas in the scene. Also, to understand how the target movement speed affects the effectiveness of \AdvTraj, for each scenario, we additionally perform 100 randomized simulations for each target movement speed group set between 0.5-2.5 m/s.\looseness=-1

We also assess the impact of the different pedestrian appearances on the attack's effectiveness against MOT algorithms with  ReID model embedded. For the same set of adversarial trajectories generated by \AdvTraj, we vary the appearances of the attacker and target with 25 different combinations of CARLA blueprints (different styles and colors of clothing). Then, the appearance-mutated simulations are replayed to ReID-enabled MOT algorithms to recalculate ASRs under each attacker-target appearance pair. The difference in appearances is calculated using the average cosine distance (same metric used by the evaluated MOT algorithms) between the attacker's and target's ReID feature vectors across all frames. The corresponding results are presented in Appendix~\ref{appearance}.\looseness=-1

\vspace{2pt}\noindent\textbf{Implementation Details.} We ran the simulation on a desktop with
i9-13900K CPU and RTX 4090 GPU.
We used the Adam optimizer~\cite{adam} with a learning rate
$\alpha=0.1$, the number of iterations $\mathtt{iter}=5$, and set the maximum
center displacement $\epsilon=0.35$ (attacker walker cannot walk faster than 3.5 m/s).\looseness=-1 

\vspace{2pt}\noindent\textbf{Real-World Experiment Setup.} We conducted real-world experiments 
to evaluate the attack effectiveness for  $(\mathtt{T2})$ adversaries. In both scenarios, the attacker only observes the positions of the MOT system's image sensor and the target walker, without knowledge of the deployed MOT system nor the ability for real-time optimization and precise motor controls. The image sensor captures $1920\times 1080$ videos at 30 frames per second, while YOLOv5 ~\cite{Jocher_YOLOv5_by_Ultralytics_2020} is used as the OD model. Wearing similar/distinct outfits than the target (black-black/black-white), the attacker walker uses one of the two black-box ID-Transfer UAMs explained in Section \ref{black-box-attack} depending on the relative starting position (Go-and-Stop if behind the target, Stop-and-Go if ahead of the target).\looseness=-1

For pedestrian surveillance, a fixed RGB sensor is mounted on the rooftop acting as the surveillance camera monitoring an outdoor terrace. The target pedestrian randomly picks a starting position and walks straight forward at a constant speed of around 1.1 M/s. We conducted the experiments 40 times for each UAM.

For the AD application, we emulate the perceived video stream of a vehicle by mounting an RGB sensor on the front hood of a Toyota Prius sedan to record footage for offline analysis. For each of the parallel and perpendicular scenarios, we perform the attack 40 times. The experiments are carried out safely in a large empty parking lot.\footnote{Black-Box \AdvTraj demo: \url{https://youtu.be/ETuJQFlxqIU}}

\begin{figure}[t]
\begin{table}[H]
\caption{
\AdvTraj white-box and black-box attack success rates in randomized simulation on different MOT models by $(\mathtt{T1})$ optimized adversary, compared with baseline ID misassignment rates when no attack is performed.\looseness=-1
}
\label{tab:white_box_asr}
\renewcommand{\arraystretch}{1}
\setlength{\tabcolsep}{6pt}
\resizebox{1\columnwidth}{!}{%
\begin{tabular}{l|m{2cm}|m{2cm}|m{2cm}|m{2cm}|m{2cm}|}
\cline{2-4}  & \textbf{Pedestrian Surveillance} & \textbf{AD-Perpendicular} & \textbf{AD-Parallel}\\ \hline
\multicolumn{1}{|c|}{White-Box} & \multicolumn{3}{|c|}{Attack Success Rate / Baseline}\\ \hline
\multicolumn{1}{|l|}{\textbf{SORT}}  & 100\% / 0\%  & 100\% / 15\% & 100\% / 6\% \\ \hline\hline
\multicolumn{1}{|c|}{Black-Box} & \multicolumn{3}{|c|}{Transfer Attack Success Rate / Baseline}\\ \hline
\multicolumn{1}{|l|}{\textbf{ByteTrack}} & 89\% / 0\% & 45\% / 16\% & 66\% / 5\% \\ \hline
\multicolumn{1}{|l|}{\textbf{OC-SORT}} & 92\% / 0\%  & 69\% / 15\% & 69\% / 3\% \\ \hline
\multicolumn{1}{|l|}{\textbf{Deep OC-SORT}} & 93\% / 0\%  & 74\% / 15\% & 80\% / 4\% \\ \hline
\multicolumn{1}{|l|}{\textbf{BoT-SORT}} & 74\% / 0\% & 78\% / 12\% & 79\% / 10\%\\ \hline
\multicolumn{1}{|l|}{\textbf{StrongSORT}} & 84\% / 0\% & 77\% / 6\% & 85\% / 2\% \\ \hline
\end{tabular}
}
\end{table}
\end{figure}

\vspace{2pt}\noindent\textbf{Baseline.} \AdvTraj poses a novel threat that has not been explored before, where the ID-Transfer differs from previous works in both attacker goal (confuse two tracked objects instead of one) and attack methodology (using physical adversarial trajectory instead of attacking OD). Therefore, to better quantify the effectiveness of our attack, we collect baseline cases for each scenario where two pedestrians have random intersecting trajectories (different speeds, directions, movement patterns etc). Specifically, for each of the pedestrian surveillance, AD-perpendicular, and AD-parallel, we collect 100 samples in CARLA (where the attacker and target have the exact same appearance for estimating an upper bound of ID misassignment rate) and 100 samples in real-world (50 wearing black/black outfit and 50 wearing black/white outfit). We evaluate the rate of ID misassignment (ID-Transfer in either direction) under these baseline cases.

\subsection{Simulation Results}\label{white_box_sim}

\vspace{2pt}\noindent\textbf{Attack Effectiveness and Transferability.} Table \ref{tab:white_box_asr} details the baseline ID misassignment rates,  white-box ASR on SORT and black-box transfer ASR on other MOT algorithms, 
where the attacker and target share the same appearance. 

The white-box attack on SORT achieved a 100\% success rate across all scenarios, emphasizing its complete vulnerability when full system knowledge is available. The black-box transfer attacks on other MOT algorithms show varying degrees of success. Notably, ByteTrack and OC-SORT show higher resilience to the transfer attack compared to the other three ReID-enabled MOT algorithms. This suggests similar appearances amplify the ID-Transfer potential for these models where their performance can degrade to be worse than motion-only MOT algorithms. Nevertheless, the transfer ASRs surpass all evaluated algorithms' baseline ID misassignment rates. This implies that even when the attacker lacks complete knowledge, the system could still face threats from black-box transfer attacks.

\vspace{2pt}\noindent\textbf{Impact of Different Target Speeds.} Figure~\ref{fig:diff_speed} shows the white-box ASRs on SORT and black-box ASRs on other MOT algorithms where the attacker and target have the same appearance but the target movement speeds are set differently. Note that the ASRs on the white-box victim algorithm SORT almost remain 100\% for different target movement speeds, except dropping slightly when the target moves at 2.5 m/s in the AD-perpendicular case. Although the white-box \AdvTraj consistently generates adversarial trajectories before successful ID-Transfer, a few unsuccessful scenarios in AD-perpendicular happened when the attacker randomly spawned at locations too far from the target to be able to close the distance within the maximum simulation duration set at 150 frames. On the other hand, the black-box ASR on the other MOT algorithms generally decreases, although mildly, as the target movement speed increases from 0.5 to 2.5 m/s.

\subsection{Real-World Experiment Results}\label{black_box_exp}

\vspace{2pt}\noindent\textbf{Attack Effectiveness.} The real-world experiment results summarized in Table~\ref{tab:black_box_asr} demonstrate the effectiveness of UAMs performed by ($\mathtt{T2}$) adversaries against various MOT algorithms under black-box settings.

StrongSORT demonstrates high robustness to the UAMs except for the AD-parallel scenario when the attacker and target have similar appearances, while other MOT algorithms are generally more susceptible in the AD-perpendicular scenario. This is because the horizontal placement of RGB sensor in this case enables the attacker to have a larger overlap with the target's bounding box (hence more flexibility in optimizing the adversarial objective). For pedestrian surveillance and AD-parallel scenarios, the vulnerabilities vary for different MOT algorithms. For example, ByteTrack is more susceptible in the pedestrian surveillance case, whereas BoT-SORT and StrongSORT are more vulnerable to AD-parallel when the two walkers' appearances are similar.

Note that the visual distinction between the attacker and target helps alleviate the attack's efficacy, as expected, in Deep OC-SORT, BoT-SORT, and StrongSORT, which have the ReID model enabled. Although BoT-SORT and SrongSORT demonstrate higher robustness in the distinct appearance cases, Deep OC-SORT still falls susceptible to the attack with ASR greater than 20\%. This suggests that the appearance matching and/or update mechanisms in Deep OC-SORT are less robust compared to BoT-SORT and StrongSORT. StrongSORT's high resilience against the adversarial trajectories in real-world experiments reflects that its incorporation of the NSA Kalman Filter (with adaptive covariance parameters) and non-linear Gaussian-smoothing interpolation during short-term occlusions may help mitigate the effect from UAMs by non-automated agents.\looseness=-1

On the other hand, since SORT, ByteTrack, and OC-SORT do not consider appearance but only use motion information for detection-tracker association, the differences in ASR between experiments with similar/different appearances are small and can be attributed to natural variations in real-world experiments.\looseness=-1

\begin{figure}[t]

\begin{table}[H]
\caption{\AdvTraj black-box attack success rates in real-world experiments by $(\mathtt{T2})$ heuristic adversary, compared with baseline ID misassignment rates without attack.\looseness=-1}
\label{tab:black_box_asr}
\renewcommand{\arraystretch}{1}
\resizebox{1\columnwidth}{!}{%
\begin{tabular}{l|m{2cm}|m{2cm}|m{2cm}|m{2cm}|}
\cline{2-4} & \textbf{Pedestrian Surveillance} &  \textbf{AD-Perpendicular} & \textbf{AD-Parallel} \\ \hline
\multicolumn{4}{|c|}{Similar Appearance - Attack Success Rate / Baseline}\\ \hline
\multicolumn{1}{|l|}{\textbf{SORT}}  & 30\% / 8\% &  37.5\% / 16\% & 30\% / 15\%  \\ \hline
\multicolumn{1}{|l|}{\textbf{ByteTrack}}  & 40\% / 12\% &  37.5\% / 16\% & 20\% / 10\% \\ \hline
\multicolumn{1}{|l|}{\textbf{OC-SORT}}  & 27.5\% / 6\% &  35\% / 10\% & 40\% / 5\% \\ \hline
\multicolumn{1}{|l|}{\textbf{Deep OC-SORT}}  & 45\% / 14\% & 30\% / 12\% & 45\% / 20\%\\ \hline
\multicolumn{1}{|l|}{\textbf{BoT-SORT}} & 6.5\% / 4\% & 37.5\% / 10\% & 20\% / 5\% \\ \hline
\multicolumn{1}{|l|}{\textbf{StrongSORT}}  & 0\% / 0\% & 0\% / 0\% & 15\% / 0\% \\ \hline\hline
\multicolumn{4}{|c|}{Distinct Appearance - Attack Success Rate / Baseline}\\ \hline
\multicolumn{1}{|l|}{\textbf{SORT}}  & 32.5\% / 6\% &  37.5\% / 16\% & 25\% / 15\%  \\ \hline
\multicolumn{1}{|l|}{\textbf{ByteTrack}}  & 30\% / 8\% &  32.5\% / 16\% & 20\% / 5\% \\ \hline
\multicolumn{1}{|l|}{\textbf{OC-SORT}}  & 32.5\% / 6\% &  40\% / 12\% & 30\% / 5\% \\ \hline
\multicolumn{1}{|l|}{\textbf{Deep OC-SORT}}  & 22.5\% / 10\% & 32.5\% / 8\% & 20\% / 10\%\\ \hline
\multicolumn{1}{|l|}{\textbf{BoT-SORT}} & 5\% / 2\% & 17.5\% / 2\% & 0\% / 0\% \\ \hline
\multicolumn{1}{|l|}{\textbf{StrongSORT}}  & 0\% / 0\% & 0\% / 0\% & 0\% / 0\% \\ \hline
\end{tabular}%
}
\end{table}
\end{figure}
\section{Discussion}\label{sec:discussion}

\vspace{2pt}\noindent\textbf{Limitations.}\label{limitation} From the evaluation results, there remains a gap in attack success rates (ASR) between the simulation and real-world experiments due to different capability levels of the $(\mathtt{T1})$ optimized and $(\mathtt{T2})$ heuristic adversaries. A successful heuristic-based attack in the physical world relies on the attack performer's dexterity and instinct in executing the maneuver. Therefore, the real-world experiments for universal adversarial maneuvers mainly serve as a proof-of-concept, where the attack success rates are calculated on the whole unedited footage to reflect the practicality of the attack. To empower a human attacker as an optimized adversary in practice, one may consider using augmented reality technology, which provides real-time localization and calculation capability, and overlays on the environment for trajectory guidance~\cite{understandingAR}. 

Another cause for the gap in ASR lies in the fact that the bounding boxes produced in the simulation are accurate, where the results of ID assignments can be attributed to the bounding-box trajectories. However, in real-world experiments, non-line-of-sight situations prevent the OD model from producing accurate bounding boxes for the (partially/completely) occluded object. To alleviate such impact, our real-world experiments were performed where the attacker was closer to the camera (blocks the target) for half of the total trials and was further away from the camera (blocked by the target) for the other half.

\vspace{2pt}\noindent\textbf{Attacker-Target Interaction.} In uncontrolled real-world scenarios, where the target could potentially react to the attacker’s actions, the attacker should carefully execute maneuvers to minimize suspicion. The black-box heuristics described in our paper are foundational for inducing ID-Transfer in MOT. However, to be more stealthy, the attacker can employ practical tricks such as running toward a walking pedestrian and then stopping to pick up a dropped wallet, or turning back as if they forgot something. Similarly, using the stop-and-go technique, while appearing unrelated to the target, can subtly influence the trajectory of the target without arousing suspicion. The key for the attacker is to perform these maneuvers naturally, reducing target reaction and thus maintaining the effectiveness of the attack.

\vspace{2pt}\noindent\textbf{Ethical Considerations.} To ensure ethical integrity, we took several precautions in designing and conducting our real-world experiments. Our experiments did not involve any participation from individuals outside the research team, and no identifiable private information was studied, analyzed, or stored. The experimental locations were carefully selected to be in safe, public spaces with minimal foot traffic, and any third parties who happened to be nearby were far from the experiment site and remained unidentifiable. 

\vspace{2pt}\noindent\textbf{Countermeasures.}\label{countermeasures} MOT systems with a single image sensor input work in the 2D plane and lack the capability to distinguish objects utilizing depth information, which may result in confusion when two detected bounding boxes overlap. Thus, incorporating robust depth information, such as through multi-sensor fusion with LiDAR or multiple cameras, could be a potential defense method. In addition, the motion prediction module of the MOT systems may be trained to adapt to adversarial trajectories against black-box attackers, specifically by adjusting the system parameters to better fit adversarial movement patterns. However, this approach incurs a trade-off between adversarial robustness and benign performance. MOT systems are typically configured for high tracking accuracy in benign scenarios, with parameters optimized to align with common movement patterns and minimize the distance between object detection results and predicted locations. Adjusting these parameters to enhance robustness against adversarial attacks may compromise the system's performance in normal situations, potentially causing frequent ID switches and other tracking errors when the system is not under attack. 
Further investigating this trade-off will be a future research direction.

Nevertheless, the linear motion assumption made by various SORT-like algorithms is inherent to the use of standard Kalman Filter as the motion prediction model. Although it strikes a balance between real-time performance and accuracy in the benign case, more expressive motion prediction models such as the Extended Kalman Filter~\cite{smith1962ekf} and the Particle Filter~\cite{particleFilter} can be adopted to address potential threats introduced by adversarial trajectories. Although StrongSORT has the highest computation overhead and slowest tracking speed among the evaluated MOT algorithms, its non-linear Gaussian-Smoothed Interpolation and NSA Kalman Filter are shown to make the model robust against the black-box UAMs in our real-world experiments against $(\mathtt{T2})$ adversary.\looseness=-1

The incorporation of ReID features as part of the association distance metrics, though introducing extra computation overhead, also helps dilute the contribution of motion information on matching decisions hence mitigating the attack. However, the ReID feature requires robust update and matching mechanisms to improve the MOT algorithms' overall robustness, since the bounding box of one object may contain information for another object when they overlap. This can also explain the performance gap between simulation and real-world experiments for ReID-enabled MOT algorithms, where the accurate bounding boxes for overlapping objects can result in confusion of their appearance (in simulation). Notably, a popular MOT algorithm DeepSORT~\cite{DeepSORT} uses appearance information only as the primary metric to perform the association (the motion information is only used to filter out unlikely matches), which is susceptible to ID-Transfer even when two tracked objects overlap in the benign case. Based on our real-world experiments, DeepSORT has 22.5\%-30\% ID-Transfer rates in baseline cases with distinct appearances and 35\%-40\% with similar appearances, and the black-box attack success rates range from 50\%-75\%. Although MOT algorithms can use both appearance and motion information to achieve better performance in benign scenarios, it still leaves the door open for the attacker to impact the system using adversarial trajectories, as motion prediction is essential in MOT  to consistently track objects especially when appearance/biometric data is unavailable or inapplicable.\looseness=-1

\section{Conclusions}
We introduce \AdvTraj, a novel physical ID-Transfer attack against MOT systems using adversarial trajectories. \AdvTraj exploits the vulnerabilities in the association phase commonly found in various state-of-the-art MOT algorithms for online trajectory generation, eliminating the need to attack the object detection model. We simulated the optimized attacker in CARLA and performed real-world experiments for the black-box heuristic attacker with application scenarios in pedestrian/vehicle surveillance and autonomous vehicles. We demonstrated the transferability of the attack across different MOT algorithms and evaluated the impact of appearance/speed differences between the attacker and the target. In the simulation, \AdvTraj produces attack success rates of up to 100\% for the white/black-box attackers, respectively. Our proposed universal adversarial maneuvers can achieve up to 45\% attack success rates by human performers in the real world.

Future work will expand our analysis to realize the white-box attack in the real world by relaxing the assumption that the attacker's surrogate OD model can always produce accurate bounding-box predictions and empowering the experimental subject to be an optimized attacker. 
In addition, we plan to extend our study to explore the impact of execution errors of adversarial trajectories on attack success rates and the possibility of other forms of ID-manipulation attacks and their end impact on autonomous systems, such as missed detection or bounding box merging, which often occur in current OD models when two objects are close in distance. Lastly, we plan to develop defenses against ID-manipulation attacks without sacrificing benign performance and ultimately develop a standardized evaluation framework for MOT system robustness against physical attacks. 

\section*{Acknowledgment}
Chenyi Wang would like to thank the Herbold Foundation for the Herbold Fellowship, which partly supported his work. 
The work of Chenyi Wang and Ming Li was supported in part by FY24 and FY25 Eighteenth Mile TRIF funding from the University of Arizona. Chenyi Wang, Ming Li, and Ryan Gerdes were supported in part by the U.S. Army Research Office (ARO) under grant W911NF-21-1-0320. The work of Z. Berkay Celik was supported in part by the National Science Foundation (NSF) under grant CNS-2144645 and grant IIS-2229876.  
We also thank our lab members: Jingcheng Li, Zhiwu Guo, and Ziqi Xu for helping with the real-world experiments.


\printbibliography

@article{understandingAR,
  title={Understanding augmented reality: Concepts and applications},
  author={Craig, Alan B},
  year={2013},
  publisher={Newnes}
}

@article{MOT17,
	title = {{MOT}16: {A} Benchmark for Multi-Object Tracking},
	shorttitle = {MOT16},
	journal = {arXiv:1603.00831},
	author = {Milan, A. and Leal-Taix\'{e}, L. and Reid, I. and Roth, S. and Schindler, K.},
	month = mar,
	year = {2016},
	note = {arXiv: 1603.00831}
}

@article{particleFilter,
author = {Hans R. K{\"u}nsch},
title = {{Particle filters}},
volume = {19},
journal = {Bernoulli},
number = {4},
publisher = {Bernoulli Society for Mathematical Statistics and Probability},
pages = {1391 -- 1403},
year = {2013}
}

@inproceedings{adam,
  author       = {Diederik P. Kingma and
                  Jimmy Ba},
  booktitle    = {ICLR},
  year         = {2015},
}

@article{hungarian,
author = {Kuhn, H. W.},
title = {The {H}ungarian method for the assignment problem},
journal = {Naval Research Logistics Quarterly},
volume = {2},
number = {1-2},
pages = {83-97},
year = {1955}
}

@misc{Jocher_YOLOv5_by_Ultralytics_2020,
author = {Jocher, Glenn},
license = {AGPL-3.0},
month = may,
title = {{YOLOv5 by Ultralytics}},
url = {https://github.com/ultralytics/yolov5},
version = {7.0},
year = {2020}
}

@InProceedings{FFAttack,
    author    = {Zhou, Tao and Ye, Qi and Luo, Wenhan and Zhang, Kaihao and Shi, Zhiguo and Chen, Jiming},
    title     = {F\&{F} Attack: Adversarial Attack against Multiple Object Trackers by Inducing False Negatives and False Positives},
    booktitle = {ICCV},
    year      = {2023},
    pages     = {4573-4583}
}

@Inproceedings{zheng2020diou,
  author    = {Zheng, Zhaohui and Wang, Ping and Liu, Wei and Li, Jinze and Ye, Rongguang and Ren, Dongwei},
  title     = {Distance-{I}o{U} Loss: Faster and Better Learning for Bounding Box Regression},
  booktitle = {AAAI},
  year      = {2020},
}

@InProceedings{meinhardt2021trackformer,
    title={Track{F}ormer: Multi-Object Tracking with Transformers},
    author={Tim Meinhardt and Alexander Kirillov and Laura Leal-Taixe and Christoph Feichtenhofer},
    year={2022},
    booktitle = {CVPR},
}

@article{centerTrack,
  title={Tracking Objects as Points},
  author={Zhou, Xingyi and Koltun, Vladlen and Kr{\"a}henb{\"u}hl, Philipp},
  journal={ECCV},
  year={2020}
}

@inproceedings{retinaTrack,
  author       = {Zhichao Lu and
                  Vivek Rathod and
                  Ronny Votel and
                  Jonathan Huang},
  title        = {Retina{T}rack: Online Single Stage Joint Detection and Tracking},
  booktitle    = {CVPR},
  pages        = {14656--14666},
  year         = {2020}
}

@inproceedings{DBLP:conf/eccv/MullerBGAG18,
  author       = {Matthias M{\"{u}}ller and
                  Adel Bibi and
                  Silvio Giancola and
                  Salman Al{-}Subaihi and
                  Bernard Ghanem},
  title        = {Tracking{N}et: {A} Large-Scale Dataset and Benchmark for Object Tracking
                  in the Wild},
  booktitle    = {ECCV},
  volume       = {11205},
  pages        = {310--327},
  year         = {2018}
}

@InProceedings{strongsort,
    author    = {Du, Yunhao and Wan, Junfeng and Zhao, Yanyun and Zhang, Binyu and Tong, Zhihang and Dong, Junhao},
    title     = {{GIAOT}racker: A Comprehensive Framework for {MCMOT} With Global Information and Optimizing Strategies in {V}is{D}rone 2021},
    booktitle = {ICCV Workshops},
    year      = {2021},
    pages     = {2809-2819}
}

@inproceedings{DBLP:conf/cvpr/FanLYCDYBXLL19,
  author       = {Heng Fan and
                  Liting Lin and
                  Fan Yang and
                  Peng Chu and
                  Ge Deng and
                  Sijia Yu and
                  Hexin Bai and
                  Yong Xu and
                  Chunyuan Liao and
                  Haibin Ling},
  title        = {LaSOT: {A} High-Quality Benchmark for Large-Scale Single Object Tracking},
  booktitle    = {CVPR},
  pages        = {5374--5383},
  year         = {2019}
}

@misc{carla-apollo-bridge,
  title = {Carla-Apollo Bridge},
  howpublished = {\url{https://github.com/guardstrikelab/carla_apollo_bridge}},
  author={Guard-Strike Lab}
}

@inproceedings{Dosovitskiy17,
  title = { {CARLA}: {An} Open Urban Driving Simulator},
  author = {Alexey Dosovitskiy and German Ros and Felipe Codevilla and Antonio Lopez and Vladlen Koltun},
  booktitle = {Annual Conference on Robot Learning},
  pages = {1--16},
  year = {2017}
}

@inproceedings{kalman1960new,
    author={Kalman, Rudolph Emil},
    title={A new approach to linear filtering and prediction problems},
    booktitle = {{ASME} Journal of Basic Engineering},
    year = {1960},
    pages = {35-45}
}

@INPROCEEDINGS{Autoware,
  author={Kato, Shinpei and Tokunaga, Shota and Maruyama, Yuya and Maeda, Seiya and Hirabayashi, Manato and Kitsukawa, Yuki and Monrroy, Abraham and Ando, Tomohito and Fujii, Yusuke and Azumi, Takuya},
  booktitle={International Conference on Cyber-Physical Systems (ICCPS)}, 
  title={Autoware on Board: Enabling Autonomous Vehicles with Embedded Systems}, 
  year={2018},
  pages={287-296}}

@INPROCEEDINGS{militaryUAV1,
  author={Luo, Xi and Zhao, Rui and Gao, Xiang},
  booktitle={International Conference on Networking, Sensing and Control (ICNSC)}, 
  title={Research on {UAV} Multi-Object Tracking Based on Deep Learning}, 
  year={2021},
  volume={1},
  pages={1-6}}

@INPROCEEDINGS{militaryUAV2,
  author={Li, Jing and Ye, Dong Hye and Chung, Timothy and Kolsch, Mathias and Wachs, Juan and Bouman, Charles},
  booktitle={International Conference on Intelligent Robots and Systems (IROS)}, 
  title={Multi-target detection and tracking from a single camera in Unmanned Aerial Vehicles ({UAV}s)}, 
  year={2016},
  pages={4992-4997}}

@article{Lin2021TraSw,
  author       = {Delv Lin and
                  Qi Chen and
                  Chengyu Zhou and
                  Kun He},
  title        = {Tra{S}w: Tracklet-Switch Adversarial Attacks against Multi-Object Tracking},
  journal      = {CoRR},
  volume       = {abs/2111.08954},
  year         = {2021},
  eprinttype    = {arXiv},
  eprint       = {2111.08954}
}

@inproceedings{Jia2020TrackerHijack,
  author       = {Yunhan Jia and
                  Yantao Lu and
                  Junjie Shen and
                  Qi Alfred Chen and
                  Hao Chan and
                  Zhenyu Zhong and
                  Tao Wei},
  title        = {Fooling Detection Alone is Not Enough: Adversarial Attack against
                  Multiple Object Tracking},
  booktitle    = {ICLR},
  year         = {2020}
}

@inproceedings{muller22TrackerHijacking,
  author       = {Raymond Muller and
                  Yanmao Man and
                  Z. Berkay Celik and
                  Ming Li and
                  Ryan M. Gerdes},
  title        = {Physical Hijacking Attacks against Object Trackers},
  booktitle    = {ACM Conference on Computer and
                  Communications Security ({CCS})},
  pages        = {2309--2322},
  year         = {2022}
}

@article{Ma2023Attack,
place = {San Diego, CA, USA}, title = {{WIP}: Towards the Practicality of the Adversarial Attack on Object Tracking in Autonomous Driving}, journal = {ISOC Symposium on Vehicle Security and Privacy}, author = {Ma, Chen and Wang, Ningfei and Chen, Qi Alfred and Shen, Chao}, year={2023}}

@misc{BaiduApollo,
  title = {Apollo: Open Source Autonomous Driving},
  howpublished = {\url{https://github.com/ApolloAuto/apollo}},
  author = {Baidu Apollo Team},
year={2017}
}

@misc{Waymo,
  title = {Waymo},
  howpublished = {\url{https://waymo.com}},
  author = {Waymo LLC}
}

@misc{OpenPilot,
  title = {Open{P}ilot},
  howpublished = {\url{https://github.com/commaai/openpilot}},
  author={comma.ai}
}

@article{elhoseny2020surveillance,
  title={Multi-object detection and tracking ({MODT}) machine learning model for real-time video surveillance systems},
  author={Elhoseny, Mohamed},
  journal={Circuits, Systems, and Signal Processing},
  volume={39},
  pages={611--630},
  year={2020},
  publisher={Springer}
}

@inproceedings{SORT,
  author={Bewley, Alex and Ge, Zongyuan and Ott, Lionel and Ramos, Fabio and Upcroft, Ben},
  booktitle={International Conference on Image Processing (ICIP)},
  title={Simple online and realtime tracking},
  year={2016},
  pages={3464-3468}
}

@inproceedings{DeepSORT,
  title={Simple Online and Realtime Tracking with a Deep Association Metric},
  author={Wojke, Nicolai and Bewley, Alex and Paulus, Dietrich},
  booktitle={International Conference on Image Processing (ICIP)},
  year={2017},
  pages={3645--3649},
}

@inproceedings{OCSORT,
  title={Observation-{C}entric {SORT}: Rethinking {SORT} for robust multi-object tracking},
  author={Cao, Jinkun and Pang, Jiangmiao and Weng, Xinshuo and Khirodkar, Rawal and Kitani, Kris},
  booktitle={CVPR},
  pages={9686--9696},
  year={2023}
}

@article{DeepOCSORT,
    title={{Deep OC-SORT}: Multi-Pedestrian Tracking by Adaptive Re-Identification}, 
    author={Maggiolino, Gerard and Ahmad, Adnan and Cao, Jinkun and Kitani, Kris},
    journal={arXiv:2302.11813},
    year={2023},
}

@article{BoTSORT,
  title={Bo{T}-{SORT}: Robust Associations Multi-Pedestrian Tracking},
  author={Aharon, Nir and Orfaig, Roy and Bobrovsky, Ben-Zion},
  journal={arXiv:2206.14651},
  year={2022}
}

@inproceedings{CVPR16ResNet,
  author       = {Kaiming He and
                  Xiangyu Zhang and
                  Shaoqing Ren and
                  Jian Sun},
  title        = {Deep Residual Learning for Image Recognition},
  booktitle    = {CVPR},
  pages        = {770--778},
  year         = {2016}
}

@inproceedings{ICCV19OSNet,
  author       = {Kaiyang Zhou and
                  Yongxin Yang and
                  Andrea Cavallaro and
                  Tao Xiang},
  title        = {Omni-Scale Feature Learning for Person Re-Identification},
  booktitle    = {ICCV},
  pages        = {3701--3711},
  year         = {2019}
}

@article{MobileNet,
  author       = {Andrew G. Howard and
                  Menglong Zhu and
                  Bo Chen and
                  Dmitry Kalenichenko and
                  Weijun Wang and
                  Tobias Weyand and
                  Marco Andreetto and
                  Hartwig Adam},
  title        = {MobileNets: Efficient Convolutional Neural Networks for Mobile Vision
                  Applications},
  journal      = {CoRR},
  volume       = {1704.04861},
  year         = {2017}
}

@inproceedings{PercepGuard,
  author       = {Yanmao Man and
                  Raymond Muller and
                  Ming Li and
                  Z. Berkay Celik and
                  Ryan M. Gerdes},
  title        = {That Person Moves Like {A} Car: Misclassification Attack Detection
                  for Autonomous Systems Using Spatiotemporal Consistency},
  booktitle    = {{USENIX} Security Symposium},
  year         = {2023}
}

@inproceedings{SzegedyZSBEGF13,
  author       = {Christian Szegedy and
                  Wojciech Zaremba and
                  Ilya Sutskever and
                  Joan Bruna and
                  Dumitru Erhan and
                  Ian J. Goodfellow and
                  Rob Fergus},
  title        = {Intriguing properties of neural networks},
  booktitle    = {ICLR},
  year         = {2014}
}

@inproceedings{DAGAttack,
  author       = {Cihang Xie and
                  Jianyu Wang and
                  Zhishuai Zhang and
                  Yuyin Zhou and
                  Lingxi Xie and
                  Alan L. Yuille},
  title        = {Adversarial Examples for Semantic Segmentation and Object Detection},
  booktitle    = {ICCV},
  pages        = {1378--1387},
  year         = {2017}
}

@inproceedings{ShapeShifter,
  author       = {Shang{-}Tse Chen and
                  Cory Cornelius and
                  Jason Martin and
                  Duen Horng Chau},
  title        = {{S}hape{S}hifter: Robust Physical Adversarial Attack on {F}aster {R-CNN}
                  Object Detector},
  booktitle    = {Machine Learning and Knowledge Discovery in Databases},
  volume       = {11051},
  pages        = {52--68},
  year         = {2018}
}

@inproceedings{DPatch,
  author       = {Xin Liu and
                  Huanrui Yang and
                  Ziwei Liu and
                  Linghao Song and
                  Yiran Chen and
                  Hai Li},
  title        = {{DPATCH:} {A}n Adversarial Patch Attack on Object Detectors},
  booktitle    = {AAAI},
  volume       = {2301},
  year         = {2019}
}

@inproceedings{TOGAttack,
  author       = {Ka Ho Chow and
                  Ling Liu and
                  Margaret Loper and
                  Juhyun Bae and
                  Mehmet Emre Gursoy and
                  Stacey Truex and
                  Wenqi Wei and
                  Yanzhao Wu},
  title        = {Adversarial Objectness Gradient Attacks in Real-time Object Detection
                  Systems},
  booktitle    = {International Conference on Trust, Privacy and Security
                  in Intelligent Systems and Applications},
  pages        = {263--272},
  year         = {2020}
}

@inproceedings{SparkAttack,
  author       = {Qing Guo and
                  Xiaofei Xie and
                  Felix Juefei{-}Xu and
                  Lei Ma and
                  Zhongguo Li and
                  Wanli Xue and
                  Wei Feng and
                  Yang Liu},
  title        = {{SPARK:} Spatial-Aware Online Incremental Attack Against Visual Tracking},
  booktitle    = {ECCV},
  volume       = {12370},
  pages        = {202--219},
  year         = {2020}
}

@inproceedings{HijackingTracker,
  author       = {Xiyu Yan and
                  Xuesong Chen and
                  Yong Jiang and
                  Shu{-}Tao Xia and
                  Yong Zhao and
                  Feng Zheng},
  title        = {Hijacking Tracker: {A} Powerful Adversarial Attack on Visual Tracking},
  booktitle    = {International Conference on Acoustics, Speech and Signal
                  Processing},
  pages        = {2897--2901},
  year         = {2020}
}

@inproceedings{UnivAttackSOT,
  author       = {Li Ding and
                  Yongwei Wang and
                  Kaiwen Yuan and
                  Minyang Jiang and
                  Ping Wang and
                  Hua Huang and
                  Z. Jane Wang},
  title        = {Towards Universal Physical Attacks on Single Object Tracking},
  booktitle    = {AAAI},
  pages        = {1236--1245},
  year         = {2021}
}

@inproceedings{Eykholt18,
  author       = {Kevin Eykholt and
                  Ivan Evtimov and
                  Earlence Fernandes and
                  Bo Li and
                  Amir Rahmati and
                  Chaowei Xiao and
                  Atul Prakash and
                  Tadayoshi Kohno and
                  Dawn Song},
  title        = {Robust Physical-World Attacks on Deep Learning Visual Classification},
  booktitle    = {CVPR},
  pages        = {1625--1634},
  year         = {2018}
}

@article{FairMOT,
  author       = {Yifu Zhang and
                  Chunyu Wang and
                  Xinggang Wang and
                  Wenjun Zeng and
                  Wenyu Liu},
  title        = {{F}air{MOT}: On the Fairness of Detection and Re-identification in Multiple
                  Object Tracking},
  journal      = {Int. J. Comput. Vis.},
  volume       = {129},
  number       = {11},
  pages        = {3069--3087},
  year         = {2021}
}

@article{ByteTrack,
  author       = {Yifu Zhang and
                  Peize Sun and
                  Yi Jiang and
                  Dongdong Yu and
                  Zehuan Yuan and
                  Ping Luo and
                  Wenyu Liu and
                  Xinggang Wang},
  title        = {{B}yte{T}rack: Multi-Object Tracking by Associating Every Detection Box},
  journal      = {CoRR},
  volume       = {abs/2110.06864},
  year         = {2021},
  eprinttype    = {arXiv},
  eprint       = {2110.06864}
}

@inproceedings{CoolingShrinking,
  author       = {Bin Yan and
                  Dong Wang and
                  Huchuan Lu and
                  Xiaoyun Yang},
  title        = {Cooling-Shrinking Attack: Blinding the Tracker With Imperceptible
                  Noises},
  booktitle    = {CVPR},
  pages        = {987--996},
  year         = {2020}
}

@inproceedings{AdvTrainingOD,
  author       = {Haichao Zhang and
                  Jianyu Wang},
  title        = {Towards Adversarially Robust Object Detection},
  booktitle    = {ICCV},
  pages        = {421--430},
  year         = {2019}
}

@inproceedings{PixelDetection,
  author       = {Syed Afaq Ali Shah and
                  Moise Bougre and
                  Naveed Akhtar and
                  Mohammed Bennamoun and
                  Liang Zhang},
  title        = {Efficient Detection of Pixel-Level Adversarial Attacks},
  booktitle    = {International Conference on Image Processing},
  pages        = {718--722},
  year         = {2020}
}

@article{JPEGCompression,
  author       = {Gintare Karolina Dziugaite and
                  Zoubin Ghahramani and
                  Daniel M. Roy},
  title        = {A study of the effect of {JPG} compression on adversarial images},
  journal      = {CoRR},
  volume       = {abs/1608.00853},
  year         = {2016}
}

@article{GaborConv,
  author       = {Abdollah Amirkhani and
                  Mohammad Parsa Karimi},
  title        = {Adversarial defenses for object detectors based on {G}abor convolutional
                  layers},
  journal      = {Vis. Comput.},
  volume       = {38},
  number       = {6},
  pages        = {1929--1944},
  year         = {2022}
}

@inproceedings{Qi_adversarial_traj,
  author       = {Qingzhao Zhang and
                  Shengtuo Hu and
                  Jiachen Sun and
                  Qi Alfred Chen and
                  Z. Morley Mao},
  title        = {On Adversarial Robustness of Trajectory Prediction for Autonomous
                  Vehicles},
  booktitle    = {CVPR},
  pages        = {15138--15147},
  year         = {2022},
}

@inproceedings {acero,
author = {Ruoyu Song and Muslum Ozgur Ozmen and Hyungsub Kim and Raymond Muller and Z. Berkay Celik and Antonio Bianchi},
title = {Discovering Adversarial Driving Maneuvers against Autonomous Vehicles},
booktitle = {USENIX Security Symposium},
year = {2023},
isbn = {978-1-939133-37-3},
pages = {2957--2974},
month = aug
}

@ARTICLE{cameraCalibration,
  author={Zhang, Z.},
  journal={IEEE Transactions on Pattern Analysis and Machine Intelligence}, 
  title={A flexible new technique for camera calibration}, 
  year={2000},
  volume={22},
  number={11},
  pages={1330-1334}}

@inproceedings {Muller_VOGUES,
author = {Raymond Muller and Yanmao Man and Ryan Gerdes and Ming Li and Jonathan Petit and Z. Berkay Celik},
title = {VOGUES: Validation of Object Guise using Estimated Components},
booktitle = {USENIX Security Symposium},
year={2024}
}

@book{smith1962ekf,
  title={Application of statistical filter theory to the optimal estimation of position and velocity on board a circumlunar vehicle},
  author={Smith, Gerald L and Schmidt, Stanley F and McGee, Leonard A},
  volume={135},
  year={1962},
  publisher={National Aeronautics and Space Administration}
}

@inproceedings{SiameseTracker,
  title={Distractor-aware Siamese Networks for Visual Object Tracking},
  author={Zhu, Zheng and Wang, Qiang and Bo, Li and Wu, Wei and Yan, Junjie and Hu, Weiming},
  booktitle={ECCV},
  year={2018}
}

\appendices

\section{}


\subsection{Weighted Bipartite Matching in MOT}\label{app:bipartite_matching}

In MOT, weighted bipartite matching is crucial for associating detected objects with their corresponding trackers across video frames. A bipartite graph consists of two disjoint sets of vertices: one representing detected object bounding boxes and the other representing bounding boxes predicted from previously tracked object trajectories. Each edge between these sets is weighted, based on the cost or distance between a detection and a tracker prediction measured by Intersection-over-Union (or its variants)~\cite{zheng2020diou}. The goal is to find a matching that minimizes the total cost to encourage correct ID assignments and consistent tracking. The Hungarian algorithm (a.k.a. Kuhn-Munkres algorithm)~\cite{hungarian} is an efficient and widely used method for solving the assignment problem in MOT.

\subsection{Details of Loss Function}\label{intuition}

\begin{figure}[ht]
    \centering
    \includegraphics[width=0.9\columnwidth]{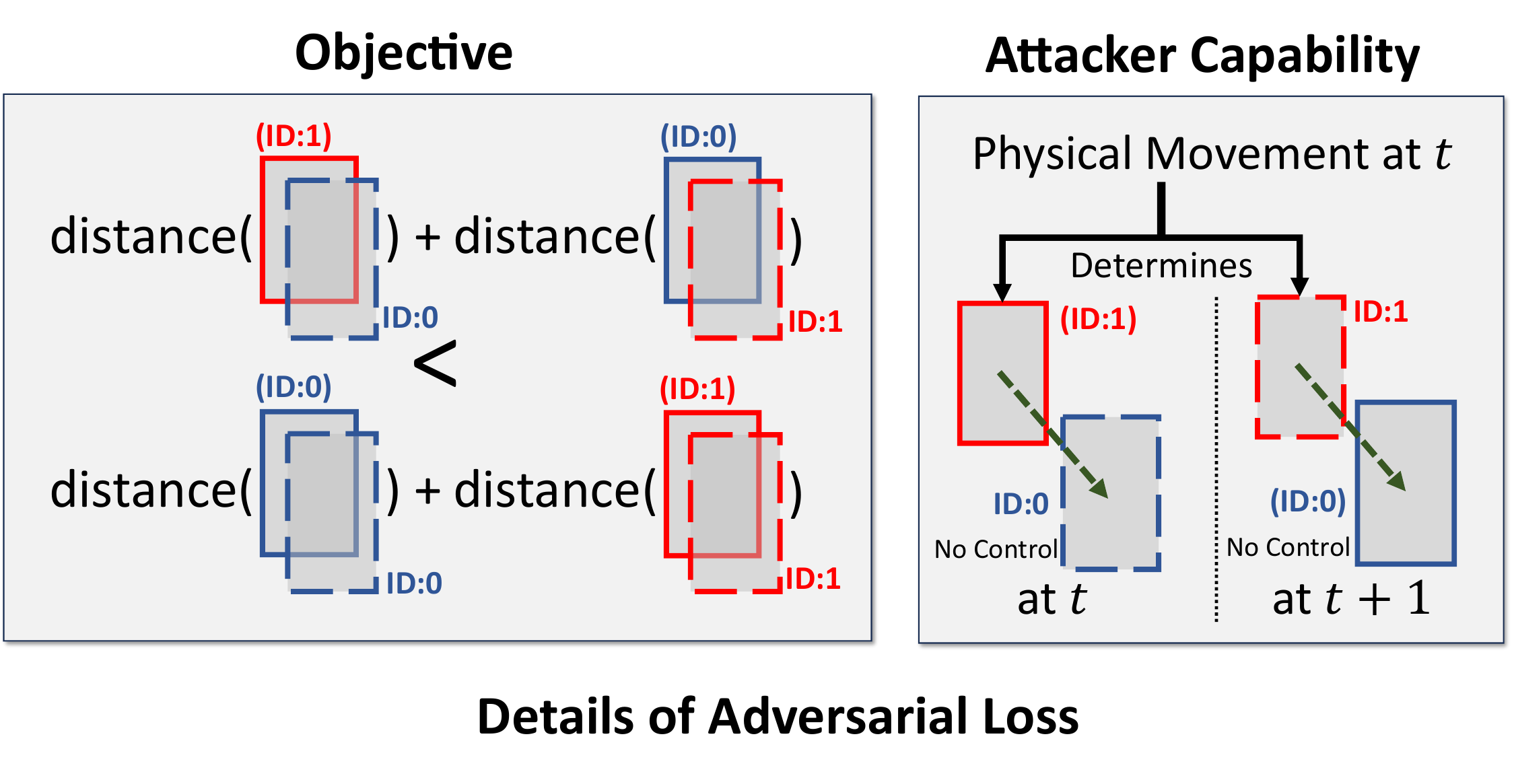}
    \caption{Illustration of the adversarial loss details. ID within parenthesis is the ground truth tracker ID the bounding box belongs to. The attacker's movement from $t-1$ to $t$ (which is optimized via gradient descent) determines its detected bounding box (solid) at $t$ and the KF predicted bounding box (dashed) at $t+1$. The attacker aims to simultaneously push its detected and KF-predicted bounding boxes closer to the target's KF-predicted and detected bounding boxes, respectively.\looseness=-1}
\label{fig:attack_objective}
\vspace{-2pt}
\end{figure}

Intuitively, minimizing Eq.~\ref{loss} produces the attacker's optimal states that achieve two goals simultaneously
: ($1$) the attacker's current states $\mathbf{d}^t$ being close to the target object's current KF predicted states $\hat{\mathbf{x}}^t_b$, which encourages the attacker to obtain the target's identity ($2$) the attacker's KF predicted states $\hat{\mathbf{x}}^{t+1}$, if updated by $\mathbf{x}^t$, being close to the target object's KF predicted states $\hat{\mathbf{x}}^{t+1}_b$ for the next time step, which promotes assigning the attacker's ID to the target. Note that the attacker's movement at any time $t$ only affects its detected states $\mathbf{x}^t$ but its corresponding next-step KF predicted states $\hat{\mathbf{x}}^{t+1}$
. Nevertheless, optimizing our loss function produces an adversarial trajectory where its history (say $t=0,...,\tau-1$) has encouraged assigning the attacker's KF predicted states to the target's detection at $\tau$. Therefore, these two goals jointly represent the best efforts at each time step of satisfying $\mathtt{C1}$:$d(\mathbf{x}_a,\hat{\mathbf{x}}_\IDb)+d(\mathbf{x}_b,\hat{\mathbf{x}}_\IDa)<d(\mathbf{x}_a,\hat{\mathbf{x}}_\IDa)+d(\mathbf{x}_b,\hat{\mathbf{x}}_\IDb)$ under our assumption that the attacker cannot influence the movement or motion prediction of the target object while the ID of the target is correctly assigned.\looseness=-1

The Intersection-Over-Union (IoU) association distance metric ($1-\text{IoU}$) used by SORT and various MOT algorithms provides no gradient when two bounding boxes have no initial overlap. To overcome this issue, we adopt d-IoU ~\cite{zheng2020diou} as the distance metric function for \AdvTraj in implementing the white-box attack:
\begin{equation}
    \mathcal{L}_{DIoU}=1-\text{IoU}+\frac{\rho^2(\mathbf{b}_1,\mathbf{b}_2)}{c^2},
\end{equation}
where $\mathbf{b}_1,\mathbf{b}_2$ denote the central points of the two bounding boxes, $\rho(\cdot)$ is the Euclidean distance, and $c$ is the diagonal length of the smallest enclosing box covering the two boxes. Note that the d-IoU distance is always lower bounded by IoU distance ($1-\text{IoU}$) and provides a non-zero gradient when two bounding boxes have no overlap yet behave similarly to IoU when overlapped. 
The evaluation of attack success rates still uses the original implementation of the MOT algorithms with default settings.\looseness=-1

\subsection{Adversarial Trajectory Visualization and Universal Adversarial Maneuver Derivation}\label{uni_adv_traj}

\begin{figure}[h]
\centering
\includegraphics[width=\columnwidth]{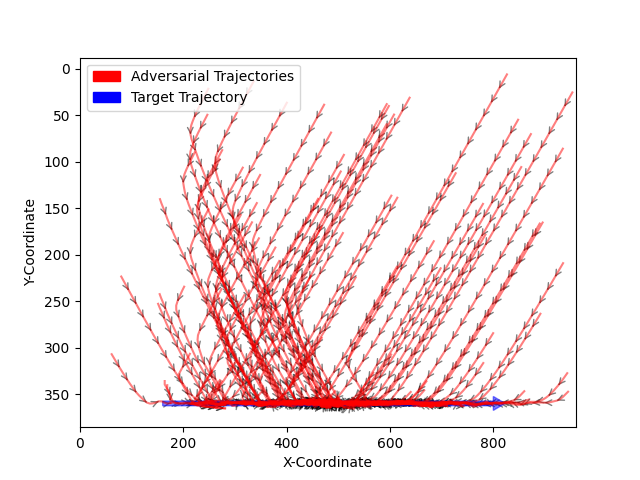}
\caption{Visualization of generated adversarial trajectories (series of bounding box centers) where the attacker is randomly spawned relative to the target.}
\label{fig:adv_traj}
\end{figure}

Although \AdvTraj is designed to work end-to-end that optimizes for the attacker's 3D physical movement, we lift the 3D-to-2D projection to generate the adversarial trajectories in 2D plane for inspecting the patterns that are agnostic to specific 3D topologies. To constrain the search space and ensure practicality in real-world scenarios, we fix the target’s movement along a straight line at a constant speed and randomly initialize the attacker's location relative to the target. We represent the trajectory using the series of bounding box centers and assume that the bounding box shape stays constant. This limits the adversarial effect to being solely dependent on the trajectory itself. The trajectory can be applied in the physical world when the camera is not parallel to the scene since the change in bounding box size (dependent on how close the object is to the image sensor) would be regarded as another `speed' dimension captured by the KF, where the same effect can be induced by moving along the adversarial trajectories. We plot the generated adversarial trajectories in Figure~\ref{fig:adv_traj}. 

\begin{figure}[t]
\centering
\begin{tabular}[t]{c}
\begin{subfigure}[t]{\columnwidth}
\begin{tikzpicture}

\definecolor{darkgray176}{RGB}{176,176,176}
\definecolor{green}{RGB}{0,128,0}
\definecolor{lightgray204}{RGB}{204,204,204}
\definecolor{orange}{RGB}{255,165,0}

\begin{axis}[
width=7cm,
height=5cm,
legend cell align={left},
legend style={
  fill opacity=0.8,
  draw opacity=1,
  text opacity=1,
  at={(0.97,0.03)},
  anchor=south east,
  draw=lightgray204
},
tick align=outside,
tick pos=left,
x grid style={darkgray176},
xlabel={Time Steps},
xmin=-7.5, xmax=157.5,
xtick style={color=black},
y grid style={darkgray176},
ylabel={Speed},
ymin=-6.5, ymax=6.5,
ytick style={color=black}
]
\addplot [very thick, red]
table {%
0 -0.150713602701823
1 0.390094757080078
2 0.938425699869792
3 1.48603947957357
4 2.02155939737956
5 2.56541697184245
6 3.00088500976562
7 3.09068171183268
8 3.14905802408854
9 3.21299362182617
10 3.29241943359375
11 3.37054061889648
12 3.42205683390299
13 3.47201029459635
14 3.50160217285156
15 3.52445602416992
16 3.55218251546224
17 3.56837336222331
18 3.59397761027018
19 3.62162526448568
20 3.63719940185547
21 3.83730951944987
22 4.03580220540365
23 4.158021291097
24 4.35763931274414
25 4.55966949462891
26 4.76077524820963
27 4.96218872070312
28 5.16485850016276
29 5.36688995361328
30 5.56924947102865
31 5.77146402994792
32 5.97460683186849
33 6.09395090738932
34 5.59005991617838
35 5.1617431640625
36 4.65815226236979
37 4.15323893229167
38 3.64964040120443
39 3.14576212565104
40 2.64090728759766
41 2.13675181070964
42 1.63241577148437
43 1.12821960449219
44 0.623087565104167
45 0
46 0
47 0
48 0
49 0
50 0
51 0
52 0
53 0
54 0
55 0
56 0
57 0
58 0
59 0
60 0
61 0
62 0
63 0
64 0
65 0
66 0
67 0
68 0
69 0
70 0
71 0
72 0
73 0
74 0
75 0
76 0
77 0
78 0
79 0
80 0
81 0
82 0
83 0
84 0
85 0
86 0
87 0
88 0
89 0
90 0
91 0
92 0
93 0
94 0
95 0
96 0
97 0
98 0
99 0
100 0
101 0
102 0
103 0
104 0
105 0
106 0
107 0
108 0
109 0
110 0
111 0
112 0
113 0
114 0
115 0
116 0
117 0
118 0
119 0
120 0
121 0
122 0
123 0
124 0
125 0
126 0
127 0
128 0
129 0
130 0
131 0
132 0
133 0
134 0
135 0
136 0
137 0
};
\addlegendentry{Attacker Speed (x)}
\addplot [very thick, green]
table {%
0 4.16963450113932
1 4.20346196492513
2 4.23512013753255
3 4.24222946166992
4 4.2507578531901
5 4.23484420776367
6 4.16904703776042
7 4.12547302246094
8 4.10265096028646
9 4.05107116699219
10 3.99793497721354
11 3.95145161946615
12 3.90614573160807
13 3.8624521891276
14 3.82111612955729
15 3.78059641520182
16 3.74125671386719
17 3.70378875732422
18 3.66963450113932
19 3.63816833496094
20 3.61164855957031
21 3.39153289794922
22 3.18847147623698
23 2.67559051513672
24 2.39890543619792
25 2.22538757324219
26 1.91137186686198
27 1.55133310953776
28 1.37602233886719
29 1.14541371663411
30 0.758341471354167
31 0.449864705403646
32 -0.00421142578125001
33 -0.0650990804036458
34 -0.157941182454427
35 0.0535736083984375
36 0.0288365681966146
37 -0.0991744995117188
38 -0.0867640177408854
39 -0.0283330281575521
40 -0.154749552408854
41 -0.225931803385417
42 -0.140780131022135
43 -0.134246826171875
44 0.0177841186523438
45 0
46 0
47 0
48 0
49 0
50 0
51 0
52 0
53 0
54 0
55 0
56 0
57 0
58 0
59 0
60 0
61 0
62 0
63 0
64 0
65 0
66 0
67 0
68 0
69 0
70 0
71 0
72 0
73 0
74 0
75 0
76 0
77 0
78 0
79 0
80 0
81 0
82 0
83 0
84 0
85 0
86 0
87 0
88 0
89 0
90 0
91 0
92 0
93 0
94 0
95 0
96 0
97 0
98 0
99 0
100 0
101 0
102 0
103 0
104 0
105 0
106 0
107 0
108 0
109 0
110 0
111 0
112 0
113 0
114 0
115 0
116 0
117 0
118 0
119 0
120 0
121 0
122 0
123 0
124 0
125 0
126 0
127 0
128 0
129 0
130 0
131 0
132 0
133 0
134 0
135 0
136 0
137 0
};
\addlegendentry{Attacker Speed (y)}
\addplot [very thick, blue]
table {%
-7.5 4.26
157.5 4.26
};
\addlegendentry{Target Speed (x)}
\end{axis}

\begin{axis}[
width=7cm,
height=5cm,
axis y line=right,
legend cell align={left},
legend style={fill opacity=0.8, draw opacity=1, text opacity=1, draw=lightgray204},
tick align=outside,
x grid style={darkgray176},
xmin=-7.5, xmax=157.5,
xtick pos=left,
xtick style={color=black},
y grid style={darkgray176},
ylabel={Distance from Attacker to Target},
ymin=-9.78237819420187, ymax=446.309799964868,
ytick pos=right,
ytick style={color=black},
yticklabel style={anchor=west}
]
\addplot [very thick,  orange, dash pattern=on 5.55pt off 2.4pt]
table {%
0 101.864849260723
1 97.3146431862647
2 92.9009388591019
3 88.6197451381739
4 84.4420292842621
5 80.3584921624693
6 76.3881497750598
7 72.5009292180275
8 68.6775042944986
9 64.9437184193337
10 61.304142432317
11 57.7584227034321
12 54.3174330565213
13 50.9917233604749
14 47.7973260903432
15 44.7541485254597
16 41.8842146268456
17 39.2171605212224
18 36.7805438696495
19 34.6065720650153
20 32.7307401355837
21 30.9596936703418
22 29.2866867001215
23 27.7913918464707
24 26.3760377239946
25 25.0190347078872
26 23.7097969879005
27 22.4279387346114
28 21.1385283858173
29 19.8174487357896
30 18.414666925816
31 16.8939553039387
32 15.2301117116578
33 13.4627634932402
34 12.1801355202311
35 11.3130516719537
36 10.9490844493922
37 11.0953688907728
38 11.7442116117924
39 12.893335201649
40 14.5497856908938
41 16.7044971024975
42 19.3691791789961
43 22.5379797374915
44 26.1923394401419
45 30.4571886499465
46 34.7414392838125
47 39.0303916655283
48 43.321402021778
49 47.6135487531658
50 51.9064070383002
51 56.1997476610608
52 60.4934336880216
53 64.7873772486109
54 69.0815188782307
55 73.375816646372
56 77.6702400002698
57 81.9647660722765
58 86.2593773595526
59 90.5540602089955
60 94.84880379552
61 99.1435994138248
62 103.438439975666
63 107.733319645536
64 112.028233571788
65 116.323177684963
66 120.618148544316
67 124.913143219494
68 129.208159198236
69 133.503194313595
70 137.79824668599
71 142.093314676648
72 146.388396849892
73 150.683491942357
74 154.97859883768
75 159.27371654556
76 163.568844184319
77 167.863980966305
78 172.159126185598
79 176.454279207606
80 180.749439460226
81 185.044606426292
82 189.339779637094
83 193.634958666807
84 197.930143127668
85 202.225332665795
86 206.520526957547
87 210.815725706351
88 215.11092863991
89 219.406135507765
90 223.701346079137
91 227.996560141022
92 232.29177749651
93 236.586997963282
94 240.882221372283
95 245.177447566528
96 249.472676400047
97 253.767907736927
98 258.063141450458
99 262.358377422367
100 266.653615542125
101 270.948855706321
102 275.244097818093
103 279.539341786625
104 283.83458752667
105 288.129834958139
106 292.425084005708
107 296.720334598473
108 301.015586669627
109 305.310840156168
110 309.606094998632
111 313.901351140847
112 318.196608529709
113 322.491867114972
114 326.787126849063
115 331.082387686899
116 335.377649585733
117 339.672912505002
118 343.968176406185
119 348.263441252683
120 352.558707009693
121 356.853973644107
122 361.149241124401
123 365.444509420548
124 369.739778503929
125 374.035048347248
126 378.330318924459
127 382.625590210696
128 386.920862182203
129 391.216134816276
130 395.511408091205
131 399.80668198622
132 404.101956481439
133 408.397231557821
134 412.692507197126
135 416.987783381868
136 421.28306009528
137 425.578337321274
};
\addlegendentry{Distance}
\end{axis}

\end{tikzpicture}
\caption{Attacker starts behind the target with an angle.}\label{pattern1}
\end{subfigure}\\
\begin{subfigure}[t]{\columnwidth}
\begin{tikzpicture}

\definecolor{darkgray176}{RGB}{176,176,176}
\definecolor{green}{RGB}{0,128,0}
\definecolor{lightgray204}{RGB}{204,204,204}
\definecolor{orange}{RGB}{255,165,0}

\begin{axis}[
width=7cm,
height=5cm,
legend cell align={left},
legend style={
  fill opacity=0.8,
  draw opacity=1,
  text opacity=1,
  at={(0.97,0.03)},
  anchor=south east,
  draw=lightgray204
},
tick align=outside,
tick pos=left,
x grid style={darkgray176},
xlabel={Time Steps},
xmin=-7.5, xmax=157.5,
xtick style={color=black},
y grid style={darkgray176},
ylabel={Speed},
ymin=-6.5, ymax=6.5,
ytick style={color=black}
]
\addplot [very thick, red]
table {%
0 -4.20144653320312
1 -4.29572550455729
2 -4.46308135986328
3 -4.63039143880208
4 -4.77323659261068
5 -4.9158681233724
6 -5.05851745605469
7 -5.20096588134766
8 -5.31827545166016
9 -5.43511454264323
10 -5.55128479003906
11 -5.64182281494141
12 -5.73205312093099
13 -5.79634094238281
14 -5.7862803141276
15 -5.77483876546224
16 -5.76194508870443
17 -5.74767049153646
18 -5.72997029622396
19 -5.70958201090495
20 -5.68606821695963
21 -5.52433522542318
22 -5.34429931640625
23 -5.19081624348958
24 -5.21450805664062
25 -5.21572113037109
26 -5.24300384521484
27 -5.20863596598307
28 -4.27721405029297
29 -4.26512908935547
30 -3.29399108886719
31 -2.32479858398437
32 -1.35962931315104
33 -0.566787719726562
34 -0.265365600585937
35 0.508595784505208
36 1.34905242919922
37 2.13203684488932
38 2.97649383544922
39 3.41681671142578
40 2.95818837483724
41 3.41681671142578
42 2.91254425048828
43 2.40694681803385
44 1.90135447184245
45 1.42911275227865
46 1.42911275227865
47 0.981821695963542
48 0.644538879394531
49 0.340568542480469
50 0
51 0
52 0
53 0
54 0
55 0
56 0
57 0
58 0
59 0
60 0
61 0
62 0
63 0
64 0
65 0
66 0
67 0
68 0
69 0
70 0
71 0
72 0
73 0
74 0
75 0
76 0
77 0
78 0
79 0
80 0
81 0
82 0
83 0
84 0
85 0
86 0
87 0
88 0
89 0
90 0
91 0
92 0
93 0
94 0
95 0
96 0
97 0
98 0
99 0
100 0
101 0
102 0
103 0
104 0
105 0
106 0
107 0
108 0
109 0
110 0
111 0
112 0
113 0
114 0
115 0
116 0
117 0
118 0
119 0
120 0
121 0
122 0
123 0
124 0
125 0
126 0
127 0
128 0
129 0
130 0
131 0
132 0
133 0
134 0
135 0
136 0
137 0
};
\addlegendentry{Attacker Speed (x)}
\addplot [very thick, green]
table {%
0 3.32588195800781
1 3.27055867513021
2 3.17116546630859
3 3.07607014973958
4 2.95898946126302
5 2.7900390625
6 2.4458974202474
7 2.14118194580078
8 1.72884623209635
9 1.35795084635417
10 1.18824005126953
11 1.07005564371745
12 0.828923543294271
13 0.512494405110677
14 0.266871134440104
15 0.251459757486979
16 0.0389073689778646
17 -0.0434519449869792
18 -0.0779469807942709
19 0.077405293782552
20 0.198557535807292
21 -0.00498199462890628
22 0.203971862792969
23 -0.219576517740885
24 -0.17486572265625
25 0.0878143310546875
26 0.0992914835611979
27 -0.140421549479167
28 -0.111183166503906
29 -0.141316731770833
30 -0.0591735839843749
31 -0.0833333333333333
32 -0.182365417480469
33 0.0746943155924479
34 -0.225474039713542
35 0.0430170694986979
36 -0.0154520670572917
37 -0.19903564453125
38 -0.162755330403646
39 -0.0959523518880208
40 -0.0899785359700521
41 -0.0959523518880208
42 -0.0789998372395833
43 -0.183057149251302
44 -0.0786361694335938
45 -0.0380325317382812
46 -0.0380325317382812
47 -0.0284652709960937
48 -0.0202255249023438
49 -0.0140864054361979
50 0
51 0
52 0
53 0
54 0
55 0
56 0
57 0
58 0
59 0
60 0
61 0
62 0
63 0
64 0
65 0
66 0
67 0
68 0
69 0
70 0
71 0
72 0
73 0
74 0
75 0
76 0
77 0
78 0
79 0
80 0
81 0
82 0
83 0
84 0
85 0
86 0
87 0
88 0
89 0
90 0
91 0
92 0
93 0
94 0
95 0
96 0
97 0
98 0
99 0
100 0
101 0
102 0
103 0
104 0
105 0
106 0
107 0
108 0
109 0
110 0
111 0
112 0
113 0
114 0
115 0
116 0
117 0
118 0
119 0
120 0
121 0
122 0
123 0
124 0
125 0
126 0
127 0
128 0
129 0
130 0
131 0
132 0
133 0
134 0
135 0
136 0
137 0
};
\addlegendentry{Attacker Speed (y)}
\addplot [very thick, blue]
table {%
-7.5 4.26
157.5 4.26
};
\addlegendentry{Target Speed (x)}
\end{axis}

\begin{axis}[
width=7cm,
height=5cm,
axis y line=right,
legend cell align={left},
legend style={fill opacity=0.8, draw opacity=1, text opacity=1, draw=lightgray204},
tick align=outside,
x grid style={darkgray176},
xmin=-7.5, xmax=157.5,
xtick pos=left,
xtick style={color=black},
y grid style={darkgray176},
ylabel={Distance from Attacker to Target},
ymin=-15.4957352968792, ymax=427.401590359992,
ytick pos=right,
ytick style={color=black},
yticklabel style={anchor=west}
]
\addplot [very thick, orange, dash pattern=on 5.55pt off 2.4pt]
table {%
0 314.032212461166
1 305.202465746367
2 296.23128821075
3 287.118734680632
4 277.89051234977
5 268.545707098662
6 259.08144531882
7 249.496220797988
8 239.814410748671
9 230.034396154424
10 220.151581702289
11 210.190102145309
12 200.147974263311
13 190.048703593629
14 179.964111623289
15 169.893011868311
16 159.835751432996
17 149.792691035447
18 139.767588676898
19 129.763171712052
20 119.781737954865
21 109.965817336401
22 100.326256584838
23 90.8462629271886
24 81.3411066380841
25 71.8300535095945
26 62.2914143921484
27 52.8441467646605
28 44.3018808391623
29 37.0807186850411
30 30.5430943446989
31 24.6674267549189
32 19.4876380027288
33 14.9184820778165
34 11.3180052235134
35 8.31626619507142
36 6.19934075310161
37 4.9762079559156
38 4.63596132388765
39 5.38019701911966
40 6.40183067293852
41 7.2853523307949
42 8.66585971059605
43 10.5621196383379
44 12.9368779323705
45 15.7317226494385
46 18.5792874161375
47 21.8646717389838
48 25.485330881709
49 29.4112487156462
50 33.6734644709084
51 37.9507947223365
52 42.2342782576409
53 46.521030962769
54 50.8097826524032
55 55.0998677364696
56 59.3908965962131
57 63.6826229276925
58 67.9748819551646
59 72.2675585441154
60 76.5605694213002
61 80.8538526422251
62 85.1473610405765
63 89.4410579877012
64 93.7349145538617
65 98.0289075529119
66 102.323018161991
67 106.617230926081
68 110.911533026511
69 115.205913734389
70 119.500363996114
71 123.794876114817
72 128.089443502557
73 132.384060485448
74 136.678722148854
75 140.973424213298
76 145.268162934136
77 149.56293501982
78 153.85773756482
79 158.15256799421
80 162.447424017594
81 166.742303590588
82 171.037204882425
83 175.332126248587
84 179.627066207562
85 183.922023421014
86 188.216996676797
87 192.511984874347
88 196.80698701206
89 201.102002176359
90 205.397029532186
91 209.692068314706
92 213.987117822039
93 218.282177408893
94 222.577246480942
95 226.872324489877
96 231.167410929019
97 235.462505329425
98 239.757607256428
99 244.05271630655
100 248.347832104744
101 252.642954301927
102 256.938082572762
103 261.233216613668
104 265.528356141025
105 269.823500889554
106 274.118650610858
107 278.413805072093
108 282.708964054766
109 287.004127353646
110 291.299294775768
111 295.594466139528
112 299.889641273859
113 304.184820017475
114 308.480002218179
115 312.775187732231
116 317.070376423769
117 321.365568164272
118 325.660762832075
119 329.955960311911
120 334.251160494499
121 338.54636327616
122 342.841568558458
123 347.136776247876
124 351.431986255512
125 355.727198496794
126 360.022412891223
127 364.317629362126
128 368.612847836436
129 372.908068244478
130 377.203290519777
131 381.498514598874
132 385.793740421158
133 390.088967928706
134 394.384197066139
135 398.679427780479
136 402.974660021024
137 407.269893739226
};
\addlegendentry{Distance}
\end{axis}

\end{tikzpicture}
\caption{Attacker starts in front of the target with an angle.}\label{pattern2}
\end{subfigure}
\end{tabular}
\caption{Two exemplar adversarial trajectories generated. Note that the acceleration of the attacker when being close the target is in the same direction as the target.\looseness=-1}
\label{fig:patterns}
\end{figure}
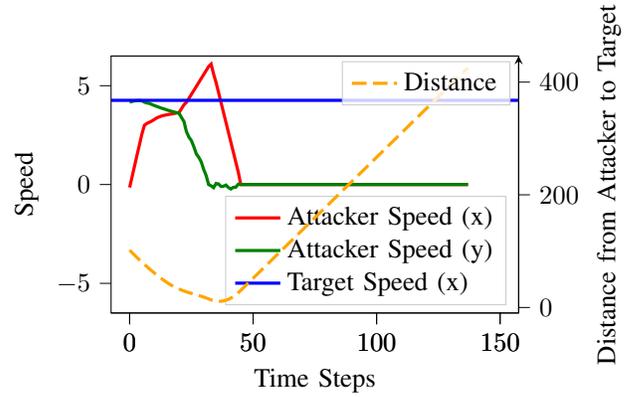
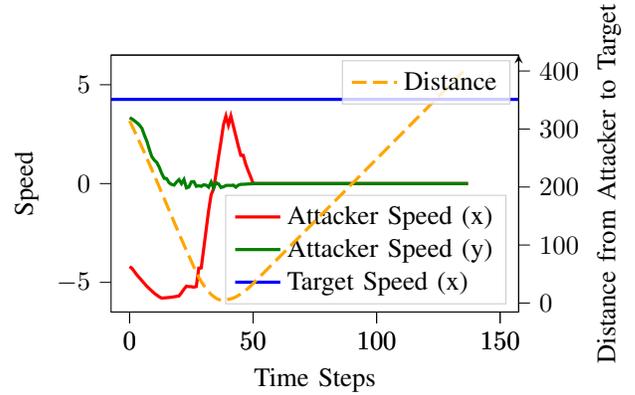

We can observe that regardless of the starting position of the attacker relative to the target, the generated trajectories first converge either to the history or projected future path of the target (corresponding to the attacker starting behind/ahead of the target), then maneuver along the same direction as the target with varying speed (non-linear movement). 
To better understand the movement pattern for each of these two cases, we randomly select an example from each case and decompose the target's velocity in both the $x$ and $y$ axes for visualization as in Figure~\ref{fig:patterns}.\looseness=-1 

In Figure~\ref{pattern1}, the attacker starts at a position behind the target. The speed on the $y$-axis gradually decreases to and stays around zero after the attacker reaches the history path of the target. At the same time, the attacker's speed on the $x$-axis increases, followed by a decrease to zero (flags the success of ID-Transfer). This motivates the design of the black-box Go-and-Stop tactic for situations where the attacker starts behind the target (even with an angle, the attacker can start by moving to the history path of the target first).

In Figure~\ref{pattern2}, the attacker starts at a position ahead of the target. The speed on the $y$-axis gradually decreases to and stays around zero after the attacker reaches the projected future path of the target. The attacker's speed on the $x$-axis was initially negative (indicating the attacker is closing the distance to the target by moving in the opposite direction in the $x$ direction). After the attacker is around the target, its speed decreases to zero (from the negative direction of $x$) and accelerates in the positive direction of $x$, followed by another decrease to zero (flags the success of ID-Transfer). However, the drastic change in moving direction may be hard to perform in the real world due to the limited maneuverability of the attacker, we design the black-box Stop-and-Go tactic for situations where the attacker starts ahead of the target (even with an angle, the attacker can start by moving to the projected history path of the target and wait for it to pass).

\subsection{Results on Vehicle Surveillance}\label{vehicle_surveillance}

\begin{table}[h]
\caption{\AdvTraj white-box attack success rates on SORT, transfer attack success rates on other MOT algorithms, and black-box attack success rates, in vehicle surveillance.}
\label{tab:vehicle_surveillance}
\renewcommand{\arraystretch}{1}
\setlength{\tabcolsep}{6pt}
\resizebox{1\columnwidth}{!}{%
\begin{tabular}{l|m{2cm}|m{2cm}|}
\cline{2-3}  & \textbf{\scriptsize CARLA Simulation} & \textbf{\scriptsize Real-World\newline Experiments}\\ \hline
\multicolumn{1}{|l|}{\textbf{\scriptsize SORT}}  & \scriptsize 83\% &\scriptsize  33.3\% \\ \hline
\multicolumn{1}{|l|}{\textbf{\scriptsize ByteTrack}} &\scriptsize  74\% & \scriptsize 26.6\% \\ \hline
\multicolumn{1}{|l|}{\textbf{\scriptsize OC-SORT}} & \scriptsize 80\% & \scriptsize 66.6\% \\ \hline
\multicolumn{1}{|l|}{\textbf{\scriptsize Deep OC-SORT}} &\scriptsize  83\% &\scriptsize  40\% \\ \hline
\multicolumn{1}{|l|}{\textbf{\scriptsize BoT-SORT}} & \scriptsize 56\% &\scriptsize  26.6\%\\ \hline
\multicolumn{1}{|l|}{\textbf{\scriptsize Strong SORT}} &\scriptsize  40\% &\scriptsize  0\%\\ \hline
\end{tabular}
}
\end{table}

For the vehicle surveillance application, the attacker and the target are two vehicles driving in the same direction in two adjacent lanes, where the camera is mounted on the roadside and monitors the roadways horizontally. The target vehicle travels forward at between 25-30 km/h. We simulate the ($\mathtt{T1}$) adversary using CARLA and perform the attack for 100 simulations with random initial positions and target driving speed in Town05. For the ($\mathtt{T2}$) adversary, we mounted an RGB sensor on a tripod in an empty open parking lot monitoring two BMW X5 SUVs (black and white) that were parallel to each other and carried out the attack 15 times. Other simulation/experiment setups are the same as in Section~\ref{sec:eval}.\looseness=-1

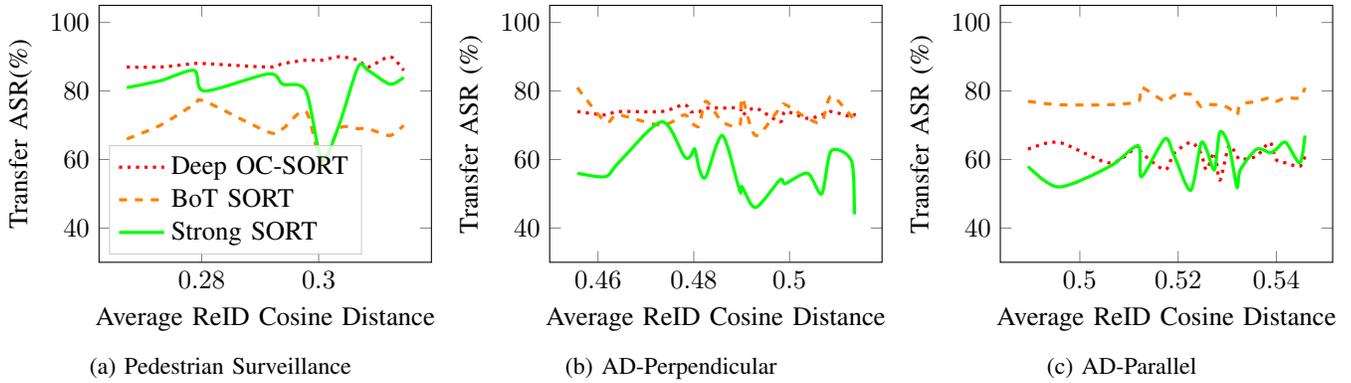
\begin{figure*}[t]
\centering
\begin{tabular}[t]{c}
\begin{subfigure}[t]{0.33\textwidth}
  \centering
\begin{tikzpicture}


\begin{axis}[
width=6cm,
height=5cm,
legend cell align={left},
legend style={fill opacity=0.8, draw opacity=1, text opacity=1, at={(0.03,0.03)}, anchor=south west, draw=white!80!black},
x grid style={white!69.0196078431373!black},
xlabel={Average ReID Cosine Distance},
y grid style={white!69.0196078431373!black},
ylabel={Transfer ASR(\%)},
ymin=30, ymax=105,
]
%
%

\addplot [very thick, dotted, red, smooth]
table {%

0.267163825757575 87
0.272931791776489 87
0.278681805613026 88
0.2804904676      88
0.291422797701425 87
0.293863977568293 88
0.297814517164260 89
0.300539014011751 89
0.303481961149104 90
0.306735764775103 89
0.308481491132445 87
0.312199443566561 90
0.314553351604082 86

}; \label{leg:deepocsort}
\addlegendentry{Deep OC-SORT}

\addplot [very thick, dashed, orange, smooth]
table {%

0.267163825757575 66
0.272931791776489 70
0.278681805613026 76
0.2804904676      77
0.291422797701425 68
0.293863977568293 69
0.297814517164260 74
0.300539014011751 59
0.303481961149104 69
0.306735764775103 69
0.308481491132445 69
0.312199443566561 67
0.314553351604082 70

}; \label{leg:botsort}
\addlegendentry{BoT SORT}

\addplot [very thick, green, smooth]
table {%

0.267163825757575 81
0.272931791776489 83
0.278681805613026 86
0.2804904676      80
0.291422797701425 85
0.293863977568293 82
0.297814517164260 80
0.300539014011751 60
0.303481961149104 70
0.306735764775103 87
0.308481491132445 86
0.312199443566561 82
0.314553351604082 84

}; \label{leg:strongsort}
\addlegendentry{Strong SORT}


\end{axis}

\end{tikzpicture}
  \caption{Pedestrian Surveillance}
\end{subfigure}
\begin{subfigure}[t]{0.33\textwidth}
  \centering
\begin{tikzpicture}


\begin{axis}[
width=6cm,
height=5cm,
legend cell align={left},
legend style={fill opacity=0.8, draw opacity=1, text opacity=1, at={(0.03,0.03)}, anchor=south west, draw=white!80!black},
x grid style={white!69.0196078431373!black},
xlabel={Average ReID Cosine Distance},
y grid style={white!69.0196078431373!black},
ylabel={Transfer ASR (\%)},
ymin=30, ymax=105,
]
%
%

\addplot [very thick, dotted, red, smooth]
table {%

0.4556861843	74
0.4615316468	73
0.464249148	74
0.4732765383	74
0.4779665773	76
0.4793275516	74
0.4802038476	74
0.4811949751	74
0.4825276693	75
0.4859714559	75
0.4895672437	75
0.4900964454	73
0.4929163025	75
0.4979653906	71
0.4991431936	74
0.5037782133	72
0.5067171904	73
0.5086950832	74
0.5129527871	72.5
0.5135334711	73.5

}; 

\addplot [very thick, dashed, orange, smooth]
table {%

0.4556861843	81
0.4615316468	71
0.464249148	73
0.4732765383	70
0.4779665773	73
0.4793275516	71
0.4802038476	70
0.4811949751	70
0.4825276693	77
0.4859714559	71
0.4895672437	70
0.4900964454	78
0.4929163025	67
0.4979653906	75
0.4991431936	76
0.5037782133	72
0.5067171904	71
0.5086950832	78.5
0.5129527871	72.5
0.5135334711	70.5

}; 

\addplot [very thick, green, smooth]
table {%

0.4556861843	56
0.4615316468	55
0.464249148	59
0.4732765383	71
0.4779665773	61
0.4793275516	61
0.4802038476	63
0.4811949751	57
0.4825276693	55
0.4859714559	67
0.4895672437	51
0.4900964454	52
0.4929163025	46
0.4979653906	54
0.4991431936	53
0.5037782133	56
0.5067171904	50
0.5086950832	62.5
0.5129527871	59.5
0.5135334711	44

}; 


\end{axis}

\end{tikzpicture}
  \caption{AD-Perpendicular}
\end{subfigure}
\begin{subfigure}[t]{0.33\textwidth}
  \centering
\begin{tikzpicture}


\begin{axis}[
width=6cm,
height=5cm,
legend cell align={left},
legend style={fill opacity=0.8, draw opacity=1, text opacity=1, at={(0.03,0.03)}, anchor=south west, draw=white!80!black},
x grid style={white!69.0196078431373!black},
xlabel={Average ReID Cosine Distance},
y grid style={white!69.0196078431373!black},
ylabel={Transfer ASR (\%)},
ymin=30, ymax=105,
]
%
%

\addplot [very thick, dotted, red, smooth]
table {%

0.489453849	63
0.4957221913	65
0.5061239084	59
0.5118690612	64
0.5125426052	62
0.5172857501	57
0.5191387515	61.66666667
0.5225342613	65
0.5246045689	62
0.5257624789	57
0.5273983652	62
0.5285047766	54
0.5303468655	64
0.531978958	61
0.5326695959	60
0.5359779436	61
0.5386759223	65
0.5399518657	60
0.5418911351	59
0.5447572654	58
0.5458167621	61

}; 

\addplot [very thick, dashed, orange, smooth]
table {%

0.489453849	77
0.4957221913	76
0.5061239084	76
0.5118690612	77
0.5125426052	81
0.5172857501	77
0.5191387515	78.66666667
0.5225342613	79
0.5246045689	75.5
0.5257624789	76
0.5273983652	76
0.5285047766	76
0.5303468655	75
0.531978958	73
0.5326695959	76
0.5359779436	77
0.5386759223	78
0.5399518657	77
0.5418911351	78
0.5447572654	78
0.5458167621	81

}; 

\addplot [very thick, green, smooth]
table {%

0.489453849	58
0.4957221913	52
0.5061239084	58
0.5118690612	64
0.5125426052	55
0.5172857501	66
0.5191387515	61.33333333
0.5225342613	51
0.5246045689	64.5
0.5257624789	63
0.5273983652	57
0.5285047766	68
0.5303468655	64
0.531978958	52
0.5326695959	57
0.5359779436	63
0.5386759223	62
0.5399518657	63
0.5418911351	65
0.5447572654	59
0.5458167621	67

}; 


\end{axis}

\end{tikzpicture}
  \caption{AD-Parallel}
\end{subfigure}
\end{tabular}
\caption{Transfer ASRs with different attacker-target appearance combinations for three applications.
	}
\label{fig:reid}
\end{figure*}

\subsection{Impact of Different Appearances}\label{appearance}

Figure \ref{fig:reid} illustrates the impact of different appearances (\eg color, outfit styles) between attacker and target walkers on the black-box ASRs against MOT algorithms with the pedestrian ReID model enabled. For the same set of generated adversarial trajectories, we mutate the outfits of the attacker and target and quantify their appearance difference by calculating the average cosine distance between the ReID feature vectors across all frames for each CARLA blueprint combination.\looseness=-1

The white-box attack conducted in the CARLA simulator demonstrates the effectiveness and transferability of \AdvTraj against various MOT algorithms using adversarial trajectories. However, due to the limited availability of walker actor blueprints offered by CARLA, the assessment of the impact of different pedestrian appearances on transfer attacks cannot extend to a wider range beyond the 25 combinations of outfits. Nevertheless, the different blueprint combinations we evaluated in simulation and the distinct appearance of attacker and target (black/white) in real-world experiments still indicate that the impact of the adversarial trajectories/maneuvers remain potent, even if the two tracked objects appear visually distinct.\looseness=-1

\subsection{End-to-End Impact on Autonomous Driving}\label{impact}

\begin{figure}[t]
\centering
\begin{tabular}[t]{c|c}
\begin{subfigure}[l]{0.4\textwidth}
    \includegraphics[width=0.35\linewidth]{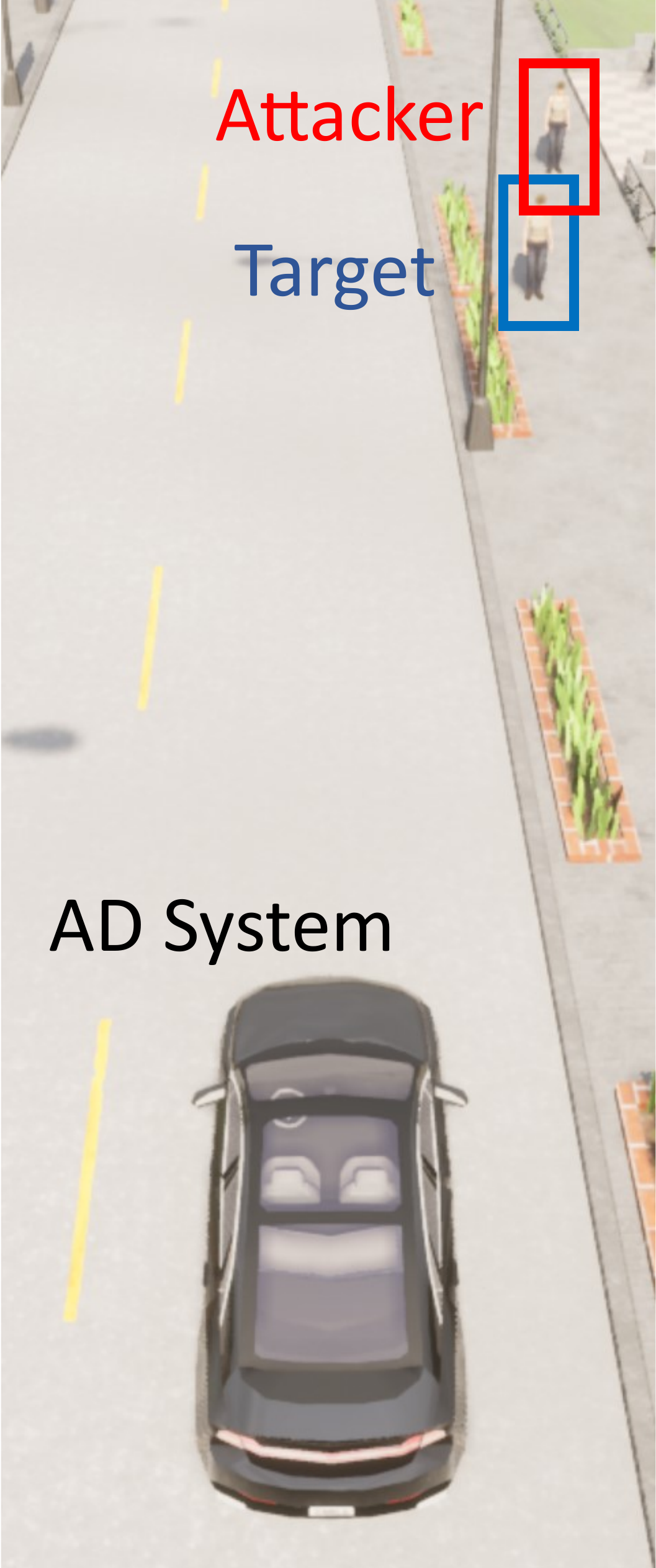}
\end{subfigure}
\hspace{-4.5cm}
     &  
        \begin{tabular}{l}
        \smallskip
        \begin{subfigure}[t]{0.3\textwidth}
            \includegraphics[width=1.0\linewidth]{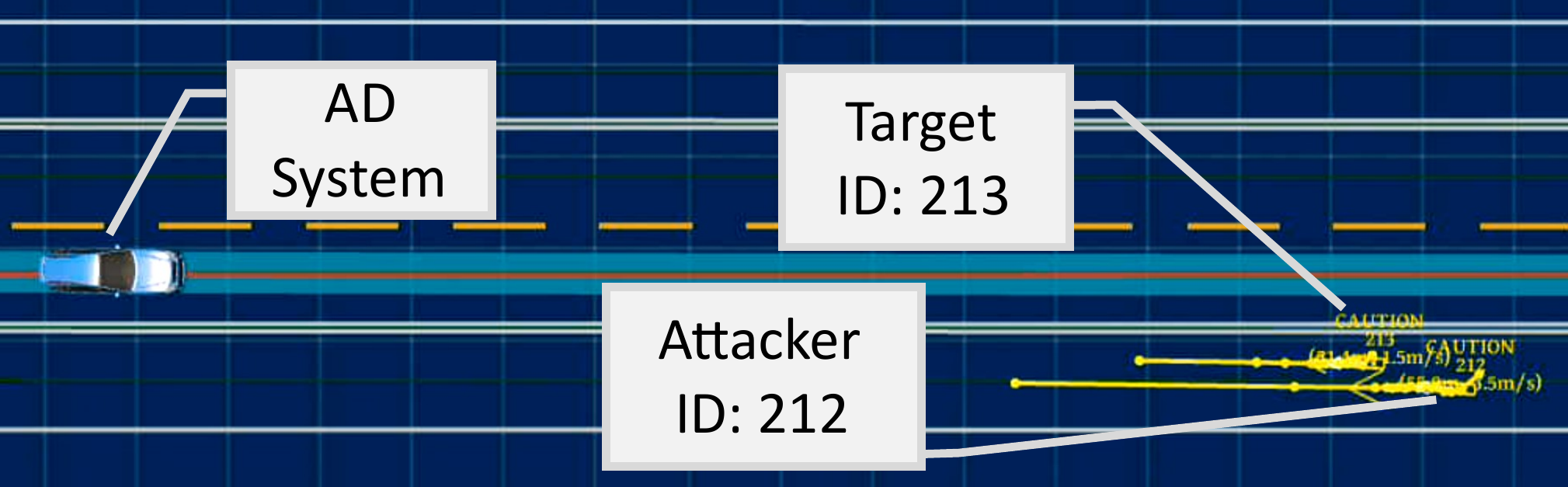}
            \caption{Before attack}\label{fig:before_attack}
        \end{subfigure}\\
        \begin{subfigure}[t]{0.3\textwidth}
            \includegraphics[width=1.0\linewidth]{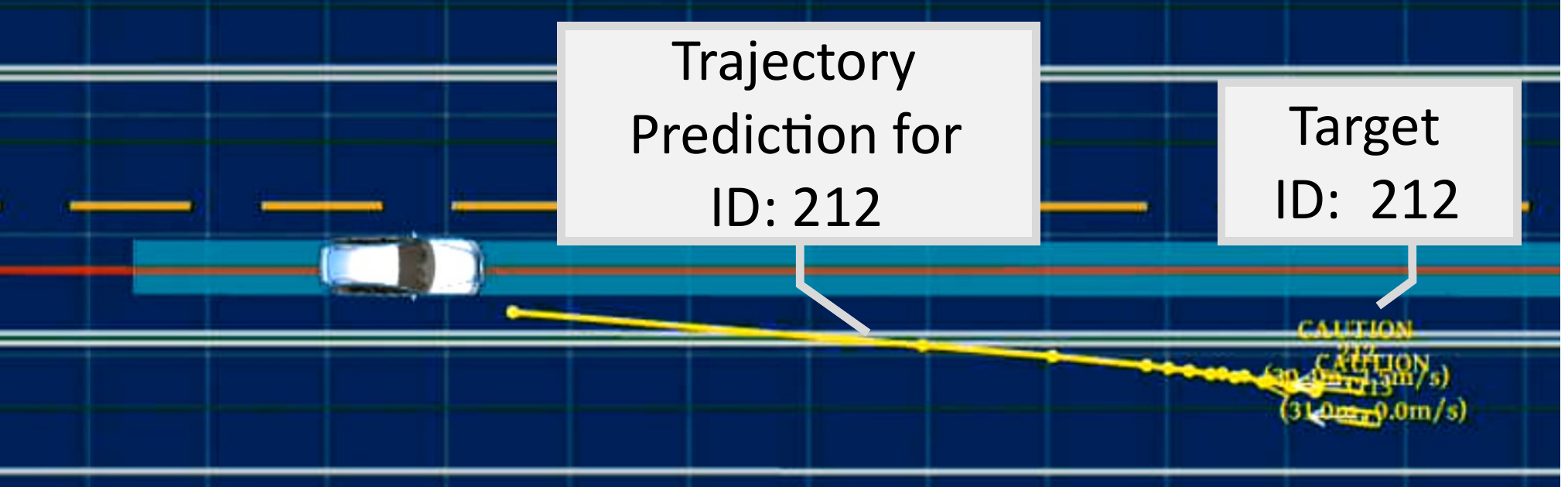}
            \caption{ID-Transfer attack}\label{fig:attack_happens}
        \end{subfigure}\\
        \begin{subfigure}[t]{0.3\textwidth}
            \includegraphics[width=1.0\linewidth]{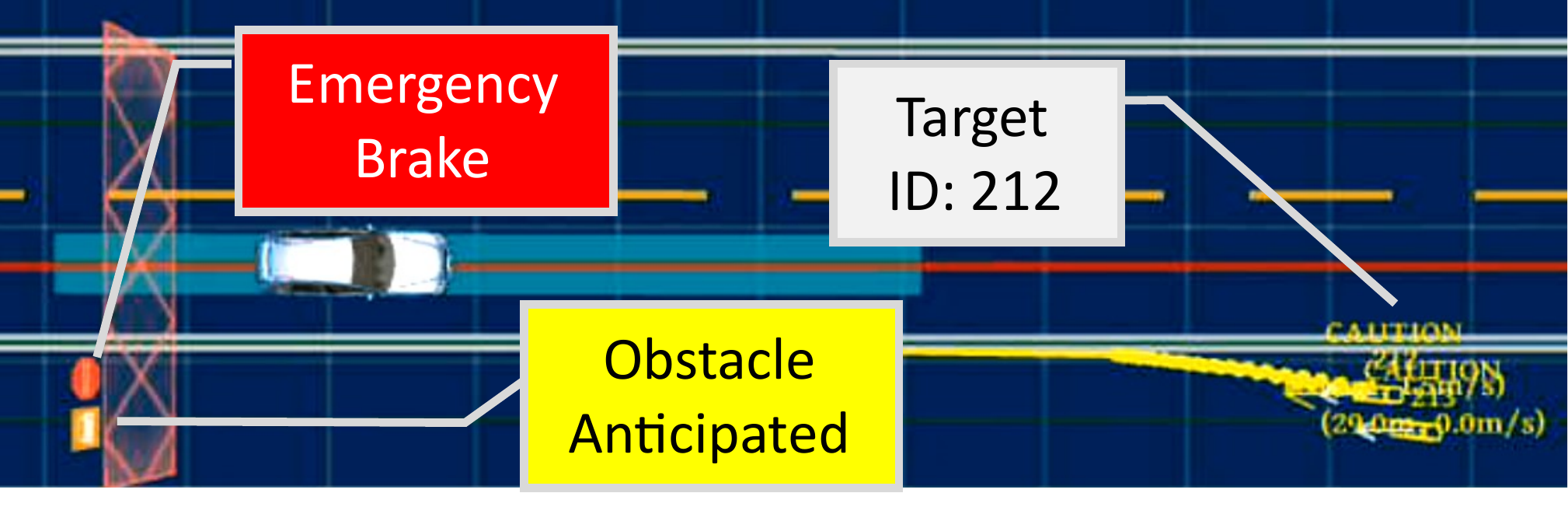}
            \caption{Emergency stop}\label{fig:av_stop}
        \end{subfigure}
        \end{tabular}\\ 
\end{tabular}
\caption{Consequence of ID-Transfer on Baidu Apollo AD system: (left) Scenario setup. (a) Before the attack. (b) The attacker walker stops walking once its ID is transferred. The history trajectory of the attacker is wrongly appended to the target's trajectory based on the assigned ID, resulting in an ``upward momentum'' that confuses the prediction module, where the predicted trajectory of the target intersects with the vehicle's roadway. (c) The AD system performs an emergency brake to avoid a wrongly anticipated collision.}
\label{fig:consequence}\vspace{-10pt}
\end{figure}

\vspace{2pt}\noindent\textbf{Case Study Setup. }As shown in Figure \ref{fig:consequence}, we construct the AD-parallel scenario in CARLA Town01 where two pedestrians on the sidewalk walk parallel to the AD system, controlled by the Baidu Apollo agent through a CARLA-Apollo bridge ~\cite{carla-apollo-bridge}. The AD system starts 200 meters ahead of the starting positions of the two pedestrians and drives autonomously under a routing request to reach a destination behind them. We visualize the trajectory prediction results of the AD on the two pedestrians and evaluate the corresponding driving decisions in both benign and adversarial cases. In the benign case, the two pedestrians walk parallel to the AD system at a constant speed of 1 m/s. In the benign case where no attack occurs, the AD system is expected to pass the two pedestrians safely, as the two pedestrians on the sidewalk should not be expected to enter the roadway. In the adversarial case, the attacker walker starts at a position behind the target and performs the black-box Go-and-Stop tactic, and its ID is transferred to the target as perceived by the AD system. The mismatch between detections and trajectory histories could lead to errors in predicting future pedestrian trajectories, which could cause unpredictable and potentially dangerous driving decisions.

\vspace{2pt}\noindent\textbf{Impact of ID-Transfer. }As shown in Figure \ref{fig:before_attack}, before the attack, the AD system produces trajectory predictions for the two pedestrians consistent with their true movement along straight paths at a constant speed. The same result is observed for the benign case where the AD runs properly and passes pedestrians without interruption. However, after the attacker performs the ID-Transfer attack, its ID is assigned to the target object, resulting in an inconsistent history update and a trajectory that ``pivots'' towards the driveway as perceived by the AD. This erroneous trajectory causes the AD system to anticipate that the target crosses the driveway as shown in Figure \ref{fig:attack_happens}. To avoid this anticipated crash, the AD decides to perform an emergency brake as seen in Figure \ref{fig:av_stop}, which leads not only to an uncomfortable experience for the passenger but also potentially to a rear-end collision by the following vehicle. Notably, the attack appears stealthy to humans, since both the attacker and target walkers are only walking along straight lines with varying speeds without explicit signs to disrupt the vehicle's operation.

\subsection{Relation to ID-Switch} 

\AdvTraj aims at transferring the ID assignments between the attacker and a targeted object, which is more stealthy than the general ID-Switch attack, where preservation of original IDs is not guaranteed. Yet, even when an attacker does not successfully transfer its ID to the target using \AdvTraj, we observe that the attempt usually results in an ID-Switch for the attacker. For example, in the black-box attack experiments for surveillance by $(\mathtt{T2})$ adversary, BoT-SORT and StrongSORT have 5\% and 0\% ID-Transfer rates between the attacker and the target, but have 80\% and 60\% ID-Switch rates for the attacker, respectively. This is because the highly non-linear adversarial movement voids the linear motion assumption made by the MOT algorithms, leading to a large distance between the tracker's linearly predicted location and the ground truth location. By employing adversarial trajectories alone, ID-Switch (the evaluated attacker goal of previous works~\cite{Jia2020TrackerHijack, Lin2021TraSw, Ma2023Attack, FFAttack} by attacking OD) can be achieved without fooling the OD model of MOT.\looseness=-1

Note that the adversarial loss function of \AdvTraj encourages swapping the IDs of the attacker and target, where both the attacker's and target's IDs are preserved yet wrongly assigned. In the case of $(\mathtt{T1})$ adversary and assuming accurate bounding box detections, swapped IDs imply successful ID-Transfer. In real-world execution, natural tracker loss may occur due to occlusions and/or detection errors (missing detections) if the target is occluded for a number of frames exceeding the threshold set by the MOT algorithm. Nevertheless, for the attacker intended to escape surveillance without raising suspicions or cause incorrect trajectory predictions, it suffices to achieve ID-Transfer where its own ID is preserved but transferred to the target.

\end{document}